\documentclass[letterpaper]{article} 

\usepackage[preprint]{neurips_2025}
\usepackage{natbib}
\usepackage{amsmath,amsfonts}
\usepackage{algorithm}
\usepackage{array}
\usepackage{textcomp}
\usepackage{stfloats}
\usepackage{url}
\usepackage{verbatim}
\usepackage{graphicx}
\usepackage{subcaption}
\usepackage{cite}
\usepackage{booktabs}
\usepackage{adjustbox}
\usepackage{placeins}
\usepackage{balance}
\usepackage[inline]{trackchanges}
\addeditor{DAW}
\usepackage{hyperref}
\usepackage{multirow}
\usepackage{xcolor}
\usepackage{xspace}
\usepackage{algcompatible}
\title{CASE: Contrastive Activation for Class-Sensitive Explanations}
\newcommand{\case}{\texttt{CASE}\xspace}
\begin{document}


\author{
  Dane Williamson\thanks{Correspondence to: \texttt{dw3zn@virginia.edu}} \\
  Department of Computer Science \\ 
  University of Virginia \\
  Charlottesville, VA, USA\\
  \texttt{dw3zn@virginia.edu}
  \and
  Yangfeng Ji \\
  Department of Computer Science \\ 
  University of Virginia \\
  Charlottesville, VA, USA\\
  \texttt{yj3fs@virginia.edu}
  \and
  Matthew Dwyer \\
  Department of Computer Science \\ 
  University of Virginia \\
  Charlottesville, VA, USA\\
  \texttt{md3cn@virginia.edu}
}

\maketitle

  \begin{abstract}
 Saliency methods are widely used to visualize which input features are deemed relevant to a model's prediction. However, their visual plausibility can obscure critical limitations. In this work, we propose a diagnostic test for class sensitivity: a method's ability to distinguish between competing class labels on the same input. Through experiments on single-label natural-image classification, we show that many widely used saliency methods produce nearly identical explanations regardless of the class label, calling into question their reliability. We find that class-insensitive behavior persists across multiple convolutional architectures and datasets, suggesting that the failure mode is not tied to a single model family or benchmark. Motivated by these findings, we introduce CASE, a contrastive explanation method that isolates features \texttt{uniquely discriminative} for the predicted class. We evaluate CASE using the proposed diagnostic and a perturbation-based fidelity test, and show that it produces consistently more class-specific explanations while maintaining competitive fidelity.
\end{abstract}
    
  \section{Introduction}
\label{sec:intro}

\par As deep learning systems become more widely deployed, explanation methods are increasingly relied upon to justify, debug, and evaluate model predictions. Activation-based approaches such as Class Activation Mapping (CAM) ~\citep{zhou2015cam} and Gradient-Weighted Class Activation Mapping (Grad-CAM) ~\citep{2016selvarajugradCAM} are among the most widely used tools for interpreting convolutional neural networks (CNNs). These methods aim to highlight features in the input that are most relevant to a particular class prediction. However, despite their popularity, such saliency methods often exhibit a critical failure mode: they produce nearly identical explanations for different class labels on the same input. \citep{hooker2019benchmark,jalwana2021cameras}. This redundancy undermines their utility in settings where distinguishing between predictions is essential, such as auditing errors, understanding misclassifications, or building trust in model behavior. \citep{alqaraawi202evaluating, lakkaraju2019faithful, lage2019evaluation,mohseni2021surveyofXAI, tjoa2021surveymedicalXAI}

\FloatBarrier



\begin{figure}[H]
    \centering
    \resizebox{0.85\columnwidth}{!}{  
    \begin{minipage}{\columnwidth}
    \centering
    \subfloat[Band Aid vs Safety Pin \label{subfig:bandaid}]{
        \includegraphics[width=\columnwidth]{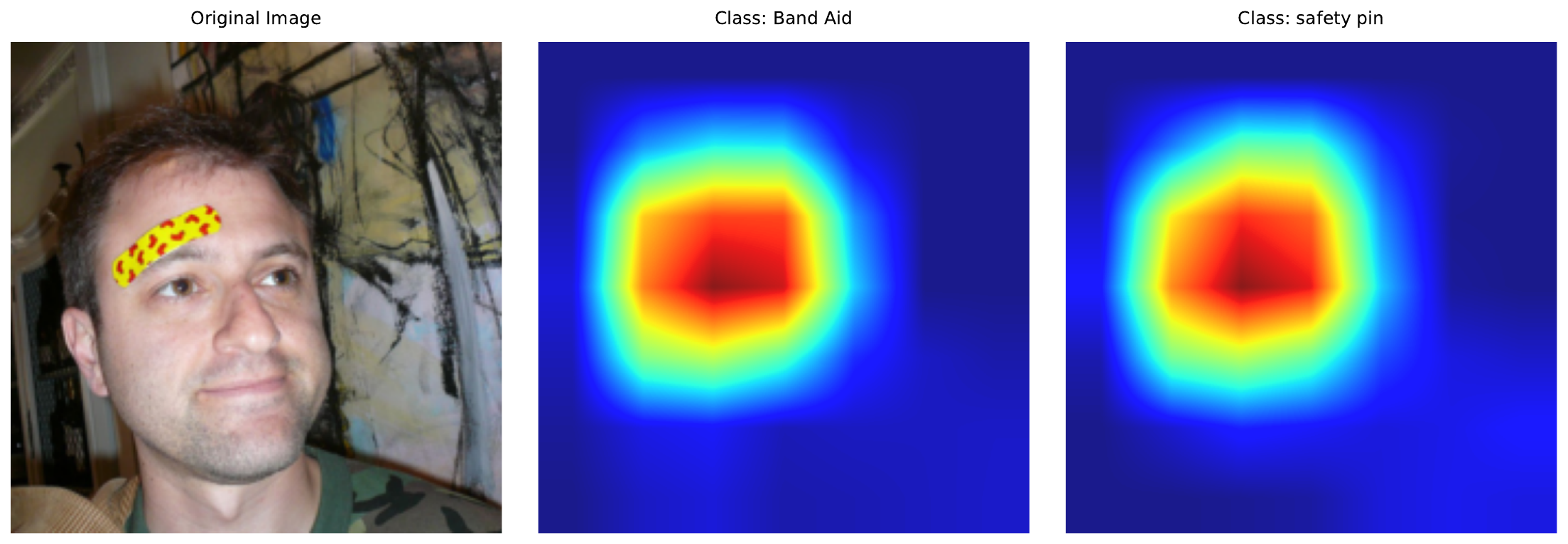}
    }
    \hspace{0.02\columnwidth}
    \subfloat[Clog vs Doormat \label{subfig:clog}]{
        \includegraphics[width=\columnwidth]{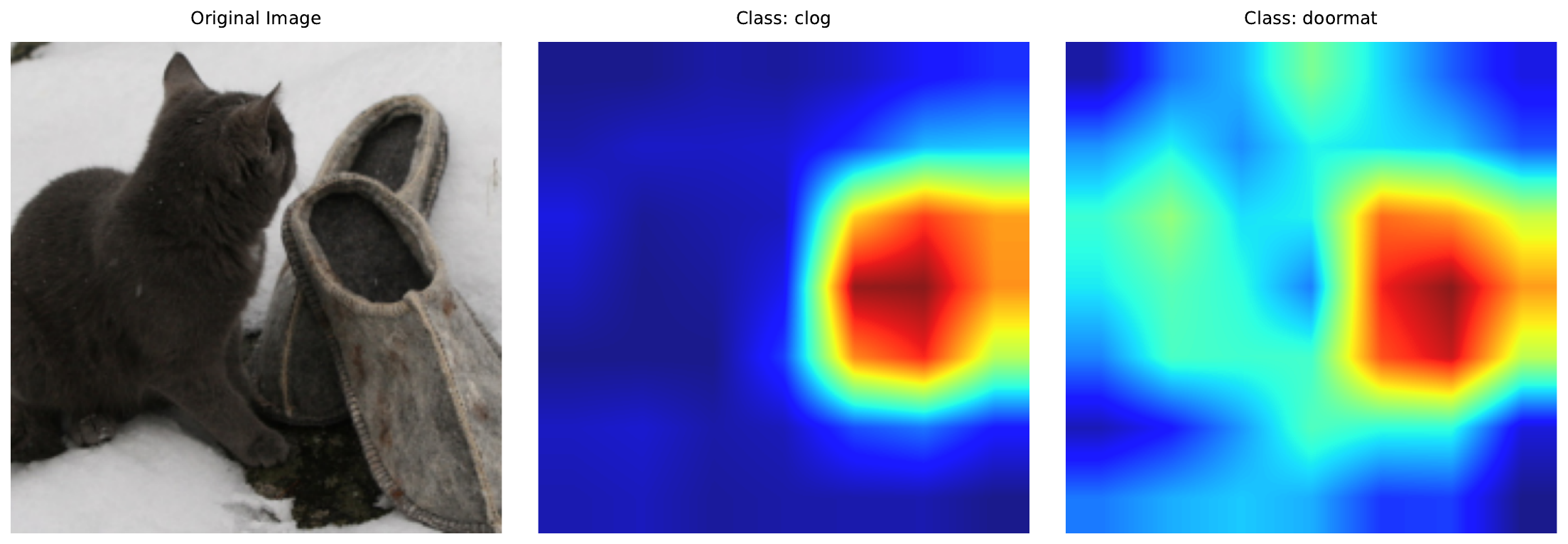}
    }

    \vspace{0.5em}

    \subfloat[Daisy vs Ant \label{subfig:daisy}]{
        \includegraphics[width=\columnwidth]{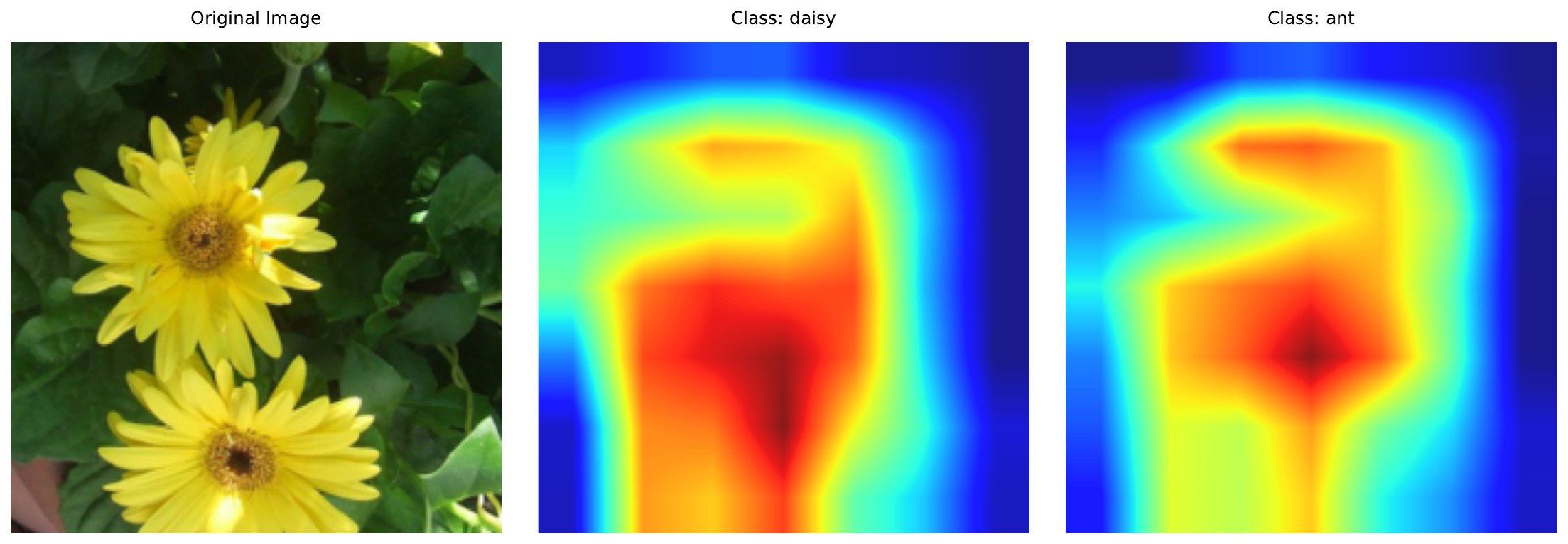}
    }
    \hspace{0.02\columnwidth}
    \subfloat[Electric Guitar vs Stage\label{subfig:guitar}]{
        \includegraphics[width=\columnwidth]{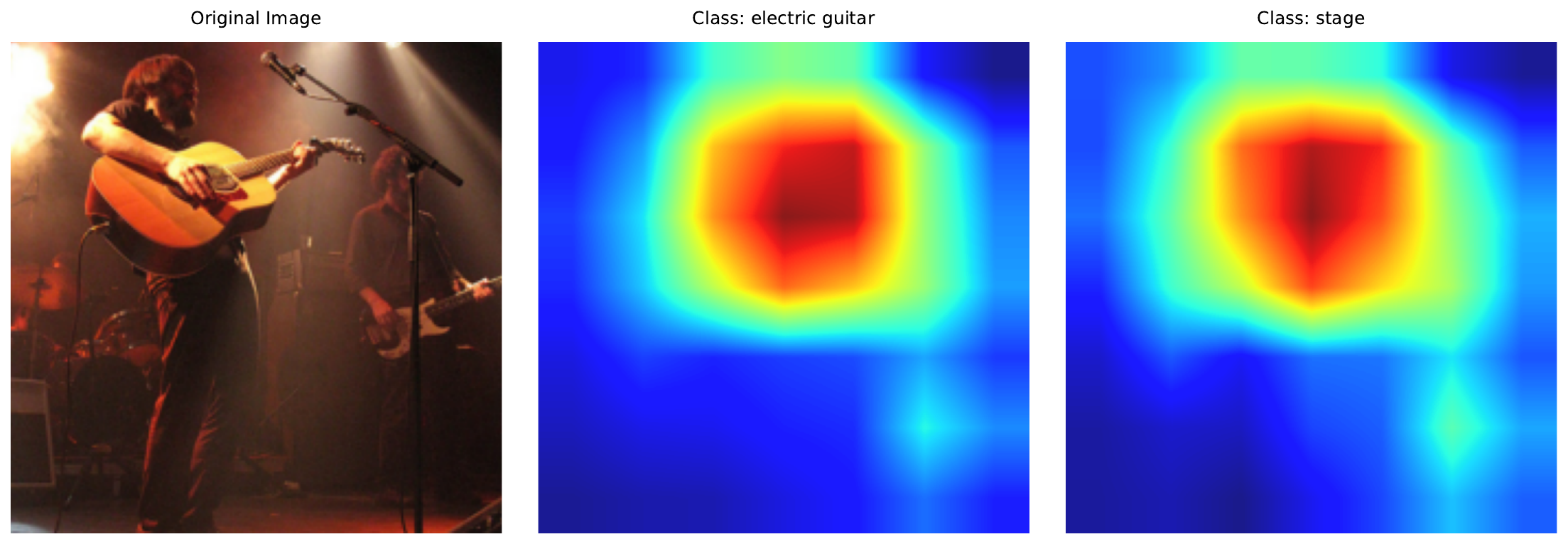}
    }
    \end{minipage}
    }
    \caption{
        Saliency maps for different class labels on the same image using Grad-CAM. 
        Despite switching the target class, the highlighted regions remain nearly identical, indicating a lack of class-specificity.
        This supports our diagnostic findings in RQ1.
    }
    \label{fig:class_insensitive_examples}
    \vspace{-0.5em}
\end{figure}

\par This concern is echoed in human-centered studies such as HIVE~\citep{kim2021hive}, which show that visual explanations can increase user trust even when the model is wrong, raising the risk that visually plausible but semantically empty saliency maps may mislead stakeholders. They also observe that existing methods often produce indistinguishable explanations for correct and incorrect predictions, suggesting a broader insensitivity to class label. In high-stakes applications, such as diagnosis or safety monitoring, this insensitivity directly compromises the value of explanations as a safeguard against error. Notably, real-world failures of saliency methods have been documented in medical imaging, where models mistakenly rely on dataset artifacts, such as text tokens or positional cues, rather than pathology features, leading to clinically misleading saliency maps. ~\citep{tjoa2021surveymedicalXAI}.

Motivated by these concerns, we focus on two essential properties that explanation methods should satisfy in classification settings:
\begin{itemize}
    \item \texttt{Class-Specificity}: Explanations should reflect features that are distinctive to the target class. Saliency maps for semantically disjoint labels (e.g., \textit{airplane} vs. \textit{barbell}) should differ more than those for related ones (e.g., \textit{terrier} vs. \textit{poodle}). \citet{bau2017dissection} demonstrate that CNN units can align with human-interpretable concepts, suggesting that class-specific representations are both learnable and quantifiable.
    
    \item \texttt{High Fidelity}: Explanations should faithfully reflect the model’s internal reasoning. This is best assessed through model-grounded metrics, such as confidence change after perturbing salient features. \citep{samek2015evaluatingLearned, petsiuk2018rise, chun2018explainCounter, chih2019sensitivity}
\end{itemize}

To illustrate this issue, \autoref{fig:class_insensitive_examples} compares saliency maps generated for the top-1 and top-2 predicted class labels on the same input. Despite the class label change, methods such as Grad-CAM produce nearly identical explanations, highlighting the same spatial regions. This visual similarity, even between semantically distinct classes, suggests that many popular saliency methods fail to isolate class-discriminative evidence.

\par We formalize this observation as a diagnostic test for class sensitivity. Specifically, we compute the top-$k$ feature overlap between saliency maps generated for different class labels on the same input. A method that highlights the same regions regardless of label fails this test. We evaluate this behavior using a one-sided Wilcoxon signed-rank test to assess whether the median agreement exceeds a lenient threshold. This analysis serves as a model- and method-agnostic check of whether a saliency method distinguishes among competing class hypotheses.

\par To address the limitation of class-insensitivity, we introduce \textbf{CASE} (Contrastive Activation for Class-Sensitive Explanations), a contrastive extension of Grad-CAM designed to highlight features that are uniquely discriminative for the target class. CASE compares neuron activations for the predicted class against those of a contrast set, reducing shared attribution and improving class specificity. Importantly, CASE requires no architectural modifications and operates on the internal activations of the trained model.

\par We evaluate CASE alongside five popular class activation map (CAM)-based saliency methods across four convolutional architectures and two single-label natural-image datasets. In our class-sensitivity diagnostic, CASE consistently produces lower top-$k$ overlap between class labels than baseline methods. In a complementary fidelity test, we measure the drop in model confidence after minimally perturbing the top-$k$ salient features. CASE achieves competitive confidence drop, indicating that its gains in class sensitivity do not generally come at the cost of attribution fidelity.

\textbf{Contributions.}
\begin{enumerate}
    \item We introduce a diagnostic test for class sensitivity that quantifies top-$k$ feature overlap between saliency maps for different class labels on the same input. Applying this test to five widely used CAM-based saliency methods across four convolutional architectures and two datasets, we find that most methods fail to distinguish between competing classes, even for semantically distinct categories.
    
    \item We propose \case, a contrastive explanation method that isolates \texttt{uniquely discriminative features} by subtracting gradient components shared with frequently confused classes. CASE consistently produces more class-specific saliency maps and maintains competitive attribution fidelity, outperforming the evaluated baselines on class sensitivity and matching them on faithfulness in most settings.
\end{enumerate}
   
  \section{Related Work}
\label{sec:related}

\subsection{Contrastive Explanations}
\label{sec:related_contrastive}
\par A number of prior methods have explored contrastive reasoning to enhance interpretability. CWOX~\citep{xie2023cwox} formulates explanation as a clustering problem over confusion sets, requiring class-specific explanations to differ from those of their nearest neighbors in a latent tree. While conceptually appealing, CWOX relies on HLTA-derived cluster structures and explicit contrast baselines, which may introduce brittleness or restrict applicability.~\citep{chen2016hlta}

\par Contrastive Layer-wise Relevance Propagation (C-LRP)~\citep{gu2018cLRP} extends the LRP \citep{bach2015lrp} framework to compare class-specific relevance maps to those of reference classes. However, C-LRP remains tied to backpropagation-based relevance signals and does not extend to activation-based methods like CAM.

\par In contrast, our approach introduces a contrastive mechanism within the CAM framework, leveraging natural competition between class logits to isolate class-discriminative evidence. We avoid the need for external baselines or learned structures, focusing instead on intrinsic class separability within the model's own prediction space.

\subsection{Saliency Methods and Class Insensitivity}
\label{sec:related_insensitivity}
\par Grad-CAM~\citep{2016selvarajugradCAM} and its variants (e.g., Grad-CAM++~\citep{chattopadhay2018gradCamplusplus}, AblationCAM~\citep{desai2020ablationCam}, ScoreCAM~\citep{wang2019scoreCam}, LayerCAM~\citep{jiang2021layerCam}) have become standard tools for visualizing model predictions. These methods produce spatial heatmaps based on either gradient or activation weighting. However, several studies have highlighted their limitations with respect to faithfulness and class sensitivity.

\par \citet{adebayo2018sanitychecks} demonstrate that certain saliency methods are invariant to randomization of model weights or labels, failing what they term \emph{"sanity checks.”}. Their work introduces the notion of \emph{model-insensitivity}, where attribution maps remain unchanged even as the model parameters are randomized. While this critique reveals concerning invariance patterns, it targets \textit{variation across models}, not \textit{across class predictions within the same model}.

\par In contrast, our work focuses on a different failure mode: \emph{class-insensitivity}. We show that many activation-based methods produce nearly identical saliency maps for different class labels\footnote{see \autoref{fig:class_insensitive_examples}}, suggesting they fail to isolate discriminative evidence. This form of insensitivity undermines the explanatory value of class-conditional attribution. While later variants of Grad-CAM improve visual smoothness or localization, they do not resolve this overlap in class-specific evidence. Our method addresses this by introducing a contrastive correction based on internal class competition, directly targeting the class-indistinct attribution problem.

\subsection{Discriminative Features and Exclusivity}
\label{sec:related_discriminative}
\par Beyond attribution intensity, several works have investigated the exclusivity of features with respect to class labels. \citet{kalibhat2023measuring} define discriminative features as those with positive gradient contributions toward a class score:

\begin{equation}
    D_c = \left\{ a \in A \;\middle|\; \frac{\partial y^c}{\partial a} > 0 \right\}
    \label{eq:discriminative}
\end{equation}

\par However, such features may overlap across classes, i.e., $\exists c \ne c' : D_c \cap D_{c'} \ne \emptyset$, leading to ambiguity. This reflects a fundamental limitation of gradient-based attribution: it reveals support, but not necessarily specificity. Our work explicitly targets this gap, focusing on isolating features that are not just correlated with the target class, but uniquely discriminative relative to plausible alternatives.

\subsection{Agreement Metrics and Evaluation of Saliency}
\label{sec:related_agreement}
\par Several metrics have been proposed to evaluate saliency maps based on consistency and selectivity. Krishna et al.~\citep{krishna2022disagreement} introduce the notion of the \emph{disagreement problem}, observing that saliency maps for different class labels often highlight overlapping regions. They propose overlap-based metrics to assess class-separability in explanations. We adopt and extend this framework by applying the Wilcoxon signed-rank test to determine whether saliency methods systematically fail to distinguish between top-1 and top-2 predictions. This offers a statistical lens on class sensitivity that complements visual inspection.

\subsection{Summary}
\label{sec:related_summary}
\par In summary, while prior work has surfaced critical limitations of saliency methods, such as insensitivity to model or data~\citep{adebayo2018sanitychecks}, or failure to isolate class-specific features ~\citep{krishna2022disagreement, kalibhat2023measuring}, few methods provide both class-distinctiveness and architectural compatibility. Our work addresses this gap by quantifying class overlap in existing methods and introducing a contrastive refinement tailored to activation-based explanations. This yields clearer, more class-sensitive visualizations and aligns explanatory focus with the model’s internal decision boundaries.

  \section[Contrastive Activation for Class-Sensitive Explanations (\case)]{Contrastive Activation for Class-Sensitive Explanations}
\label{sec:CASE}

\subsection{Overview}
\par
\begin{figure}[t]
    \centering
    \includegraphics[width=\columnwidth]{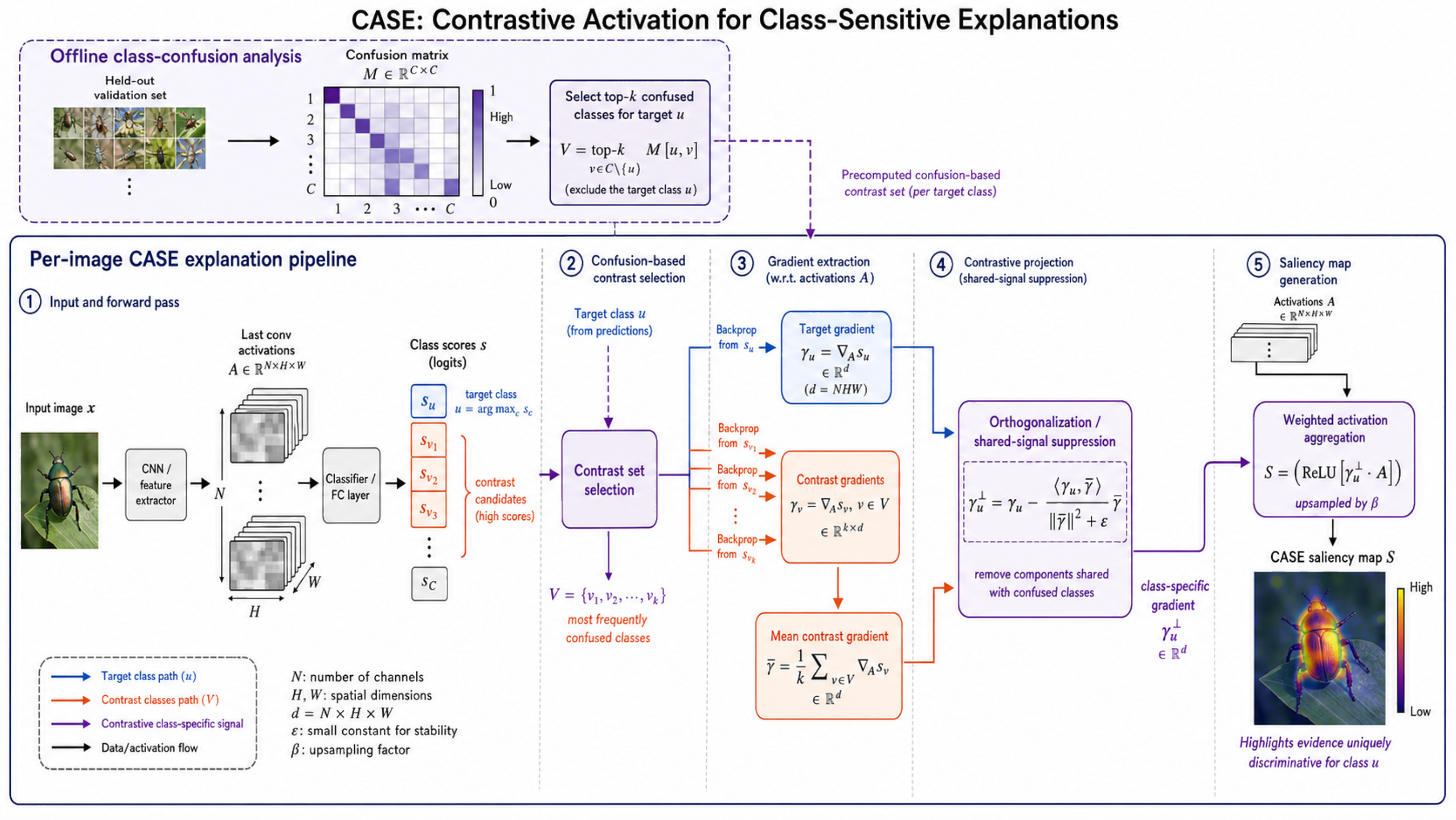}
    \caption{
    Overview of the \texttt{CASE} algorithm. Gradients are backpropagated for the target class  and contrast classes. The contrast gradients are averaged, and the target gradient is projected onto this average. The projection is then subtracted from the original target gradient to isolate a class-specific signal. The resulting orthogonal component is used to compute the weights for the final saliency map}
    \label{fig:case_overview}
\end{figure}

 We introduce \case (Contrastive Activation for Class-Sensitive Explanations), a gradient-based explanation method designed to improve class-specificity by suppressing shared attribution between frequently confused class labels. While existing saliency methods highlight features that contribute to a class prediction, they often fail to distinguish between closely related classes, yielding nearly identical explanations across labels. \case addresses this by explicitly removing gradient components shared with a contrast set, classes that the model confuses with the target, thereby isolating uniquely discriminative evidence. The full procedure is detailed in Algorithm~\ref{alg:confusion_contrastive_localization} and visualized in \autoref{fig:case_overview}.

\subsection{Contrastive Gradients}
\par Prior work has observed that many saliency methods produce overlapping explanations across distinct class labels~\citep{krishna2022disagreement}, limiting their ability to convey what distinguishes one class from another. In \case, we begin by computing the gradient of the class score $ s_u $ with respect to the activation map $ A $ for the target class $u $. We then suppress the components of this gradient that are aligned with an average gradient vector derived from a contrast set, i.e., the classes most frequently confused with $ u $. This orthogonalization removes shared attribution directions and emphasizes features that are uniquely predictive of $ u $.

\subsection{Confusion-Based Contrast Set Selection}
\label{sim_met}
To identify an appropriate contrast set, \case uses the model’s confusion matrix $ M \in \mathbb{R}^{C \times C} $, computed on a held-out validation set. For a given class $ u $, we extract the top-$ k$ most confused alternatives:

\begin{equation}
    \label{eq:contrast_class}
    \mathcal{V} = \arg\max_{v \in \mathcal{C} \setminus \{u\}}^{k} M[u, v]
\end{equation}

\par This behaviorally grounded approach ensures that the contrast set $ \mathcal{V} $ consists of labels that the model empirically struggles to distinguish from $ u $, rather than relying on semantic priors or embedding similarity. The mean gradient across this set is used to define a shared attribution direction to suppress.

\subsection{Uniquely Discriminative Features}
\label{sec:udf}
Let \( D_u \) denote the set of features positively contributing to class $ u $, and let $ D_v $ represent those for a contrast class $ v \in \mathcal{V} $. The goal of \case is to approximate the uniquely discriminative feature set:

\begin{equation}
    \label{eq:udf}
    U_u^{\mathcal{V}} = D_u \setminus \bigcup_{v \in \mathcal{V}} D_v
\end{equation}

\par By suppressing attribution that overlaps with the contrast set, \case highlights only those features that are uniquely predictive of the target class. For example, distinguishing \textit{tiger shark} from \textit{great white shark} involves removing marine features common to both, while retaining subtle class-specific traits.

                  
      
      
      
      
      

\begin{algorithm}[t]
  \caption{CASE with Confusion-Based Contrast}
  \label{alg:confusion_contrastive_localization}
  \begin{algorithmic}[1]
      \STATE \textbf{Input:} Bottleneck activation maps $A \in \mathbb{R}^{N \times H \times W}$; 
                  Confusion matrix $M \in \mathbb{R}^{C \times C}$;
                  Target class $u \in \mathcal{C}$;
                  Number of contrast classes $k$;
                  Upsampling factor $\beta$
                  
      \STATE \textbf{Output:} Saliency map $S \in \mathbb{R}^{H \times W}$
      
      \STATE Compute target gradient: $\gamma_u = \nabla_A s_u$
      
      \STATE Identify top-k confused classes: 
      \[
        \mathcal{V} = \arg\max_{v \in \mathcal{C} \setminus \{u\}}^{k} M[u, v]
      \]
      
      \STATE Initialize contrast gradient accumulator: $\bar{\gamma} = 0$
      
      \FOR{each $v \in \mathcal{V}$}
          \STATE $\gamma_v = \nabla_A s_v$
          \STATE $\bar{\gamma} \mathrel{+}= \gamma_v$
      \ENDFOR
      
      \STATE $\bar{\gamma} \mathrel{/}= k$ \hfill // Mean contrast gradient
      
      \STATE Project $\gamma_u$ onto orthogonal complement of $\bar{\gamma}$:
      \[
        \gamma^\perp_u = \gamma_u - \frac{\langle \gamma_u, \bar{\gamma} \rangle}{\|\bar{\gamma}\|^2 + \epsilon} \cdot \bar{\gamma}
      \]
      
      \STATE Compute saliency: 
      \[
        S = \left( \max \left\{ \gamma^\perp_u \cdot A,\ 0 \right\} \right) \uparrow \beta
      \]
  \end{algorithmic}
\end{algorithm}

\subsection{Saliency Map Generation}
To construct the final saliency map, \case computes the orthogonal projection of the target gradient $ \gamma_u = \nabla_A s_u $ against the average contrast gradient $ \bar{\gamma} $, where:

\begin{equation}
    \label{eq:contrast_gradient}
    \bar{\gamma} = \frac{1}{k} \sum_{v \in \mathcal{V}} \nabla_A s_v
\end{equation}

The projection is given by:

\begin{equation}
    \label{eq:projection_gradient}
    \gamma_u^\perp = \gamma_u - \frac{\langle \gamma_u, \bar{\gamma} \rangle}{\|\bar{\gamma}\|^2 + \epsilon} \cdot \bar{\gamma}
\end{equation}

\par This removes shared directional components between $ \gamma_u $ and the contrast set. The resulting class-specific saliency map is computed via a weighted sum over the activation maps and passed through a ReLU and upsampling step:

\begin{equation}
    \label{eq:case_map}
    S = \left( \operatorname{ReLU} \left[ \gamma_u^\perp \cdot A \right] \right) \uparrow \beta
\end{equation}

\par Here, $ A \in \mathbb{R}^{N \times H \times W} $ denotes the final convolutional activation tensor, where $N$ is the number of channels, and $\uparrow \beta $ indicates bilinear upsampling to input resolution. This formulation yields saliency maps that emphasize evidence uniquely associated with the predicted class and suppress attribution that generalizes across visually similar alternatives.

\subsubsection{Degenerate residual check.}
\case removes contrast-shared class evidence by projecting the target
class gradient $\gamma_u$ away from the mean contrast gradient
$\bar{\gamma}$. This projection has a potential degenerate case: if
$\gamma_u$ is nearly collinear with $\bar{\gamma}$, then the orthogonal
residual
\begin{equation}
\gamma_u^\perp =
\gamma_u -
\frac{\langle \gamma_u,\bar{\gamma}\rangle}
     {\|\bar{\gamma}\|_2^2+\epsilon}\bar{\gamma}
\end{equation}
can approach zero, producing a saliency map with little or no class-specific
signal.

To quantify this edge case, for each architecture, we sampled 500
validation candidates with a fixed seed and retained examples classified
correctly by the corresponding model, matching the evaluation regime used for
the reported class-sensitivity and fidelity analyses. RQ1 generated one \case
map for each of the top-1 and top-2 predicted classes of each retained image,
while RQ2 generated one \case map for the ground-truth class of each retained
image.

For every generated \case map, we recorded the normalized residual magnitude

\begin{equation}
r =
\frac{\|\gamma_u^\perp\|_2}
     {\|\gamma_u\|_2+\epsilon},
\qquad \epsilon = 10^{-8},
\end{equation}
along with the selected contrast classes, model/layer configuration, sample
ID, and post-ReLU map statistics, including maximum value, mean value, and
fraction of nonzero pixels. Before inspecting the results, we defined a
near-zero residual as $r < 10^{-6}$ and performed threshold-sensitivity checks
at $10^{-4}$ and $10^{-8}$.

As shown in \autoref{tab:case-near-zero-residuals}, we observed no near-zero
\case residuals in the evaluation set: $0/4774$ maps met the primary
$r < 10^{-6}$ criterion, and the count remained zero under both sensitivity
thresholds. Since no examples were flagged, the inclusive analysis and the
exclusion-sensitivity analysis were identical: excluding near-zero residual
maps did not change any RQ1 significance decision or any RQ2 confidence-drop
comparison direction or significance decision. Under the pre-specified
criterion, near-zero target/contrast gradient alignment is therefore a
negligible edge case in our evaluation and does not affect the reported
conclusions.

\begin{table}[t]
\centering
\caption{Prevalence of near-zero CASE residuals under the primary threshold
and two sensitivity thresholds. RQ1 generates two CASE maps per evaluated
image pair, while RQ2 generates one CASE map per evaluated image.}
\label{tab:case-near-zero-residuals}
\begin{tabular}{lrrrr}
\toprule
Threshold & RQ1 maps & RQ2 maps & Total maps & Near-zero \% \\
\midrule
$10^{-4}$ & $0/3182$ & $0/1592$ & $0/4774$ & $0.0000$ \\
$10^{-6}$ & $0/3182$ & $0/1592$ & $0/4774$ & $0.0000$ \\
$10^{-8}$ & $0/3182$ & $0/1592$ & $0/4774$ & $0.0000$ \\
\bottomrule
\end{tabular}
\end{table}

  \section{Experiments}
\label{sec:experiments}

\par This section presents our empirical evaluation of saliency method performance with a focus on class sensitivity and explanation faithfulness. We structure our analysis around two guiding research questions.

\subsection{Research Questions}

\textbf{RQ1:} \textbf{\textit{How does \case compare to existing saliency methods in isolating class-specific evidence across architectures?}}

\textbf{RQ2:} \textbf{\textit{Can \case improve class sensitivity without compromising explanation fidelity?}}

\subsection{RQ1: Class Sensitivity}
\label{sec:experiments_rq_one}
\par RQ1 aims to determine whether saliency methods meaningfully distinguish between class labels when generating explanations for the same input. For each image, we generate saliency maps using the top-1 and top-2 predicted class labels and compute the overlap between their top-$k$ most salient pixels. In the main experiments, $k$ for this diagnostic is fixed to the top 5\% of pixels, following the common practice of evaluating the highest-attribution region while avoiding near-empty masks. High overlap would indicate that a method tends to highlight the same regions regardless of class label, suggesting poor class sensitivity.

\par To quantify this behavior, we use the feature agreement metric proposed by \citep{krishna2022disagreement}, which computes the proportion of shared features among the top-$k$ salient pixels:
\begin{equation}
    \label{eq:feature_agreement}
    F(E, E', k) =  \frac{\lvert \{ f : f \in t(E,E',k) \} \rvert}{k}
\end{equation}
where $E$ and $E'$ are saliency maps for two different class labels, and $t(E, E', k)$ returns the intersection of the top-$k$ salient features for both.

\par We compute this score across a large sample of validation images for each method. A one-sided Wilcoxon signed-rank test is then applied to assess whether the observed agreement scores are significantly lower than would be expected under class-insensitive behavior.

\par The null hypothesis is that the median agreement between saliency maps for the top-1 and top-2 class labels is greater than or equal to 50\%. A statistically significant rejection of this hypothesis indicates that the method produces class-distinct explanations.

\begin{equation}
    \label{eq:rq_one_null_hypothesis}
    \begin{array}{rl}
        H_0: & \text{Median}\big(F(E, E', k)\big) \geq 0.50 \\
        H_1: & \text{Median}\big(F(E, E', k)\big) < 0.50
    \end{array}
\end{equation}

\subsection{RQ2: Explanation Faithfulness}
\label{sec:experiments_rq_two}
\par RQ2 evaluates whether \case preserves competitive faithfulness relative to baseline methods while improving class sensitivity. We define explanation faithfulness as the degree to which the most salient regions identified by a method correspond to regions truly influential to the model’s prediction.

\par To measure this, we ablate the top-$k$ most salient regions identified by each method and observe the resulting drop in classification confidence for the predicted class. We use the same top-5\% mask size as in RQ1 so that class-sensitivity and faithfulness results are evaluated at a matched attribution budget. Larger confidence drops indicate that the identified regions were more influential to the model’s decision, and therefore reflect greater faithfulness.

\par We compute the confidence drop for each method across the same set of evaluation samples and apply a paired t-test to compare \case against each baseline. This enables a controlled comparison of attribution fidelity while accounting for variability across inputs.

\par The null hypothesis is that the mean confidence drop from ablating \case's top-$k$ salient regions is less than or equal to the mean drop from the baseline method. Rejection of this hypothesis indicates that \case produces significantly larger confidence drops than the corresponding baseline; failure to reject indicates that \case is not statistically superior under this perturbation metric.

\begin{equation}
    \label{eq:rq_two_null_hypothesis}
    \begin{array}{rl}
        H_0: & \text{Mean}\big(D_{\case}\big) \leq \text{Mean}\big(D_{\text{baseline}}\big) \\
        H_1: & \text{Mean}\big(D_{\case}\big) > \text{Mean}\big(D_{\text{baseline}}\big)
    \end{array}
\end{equation}

\subsection{Models, Datasets, and Methods}
\label{sec:experiments_configs}
\par Our evaluation includes four pre-trained image classification models: ResNet-50, VGG-19, DenseNet-201, and ConvNeXt-Large~\citep{he2015resnet50, liu2022convNext, huang2016denseNet}. Models are evaluated on two standard datasets: ImageNet~\citep{deng2009imageNet}, which offers broad category diversity across 1,000 classes, and CIFAR-100~\citep{krizhevsky2009cifar100}, which contains 100 object categories drawn from 20 coarse-grained superclasses. CIFAR-100 consists of low-resolution natural images ($32 \times 32$) covering a wide range of everyday concepts such as animals, vehicles, and household objects.

\par For each model and dataset, we randomly sample 1000 validation images, retaining only those for which the top-1 prediction is correct. All models are evaluated using publicly available pretrained model without fine-tuning. \footnote{Unless otherwise specified}

\par We compare \case against five representative activation-based CAM-family saliency methods: Grad-CAM, Grad-CAM++, Score-CAM, Ablation-CAM, and LayerCAM. Each method is applied using standard configurations with a shared backbone to ensure consistency. All evaluation procedures are applied identically across methods. We focus on CAM-family baselines because \case operates in the same activation-map attribution setting, making the comparison direct in terms of model inputs, attribution layer, saliency-map format, and perturbation protocol. The contrast set size for \case is fixed to three confused classes in the main experiments; sensitivity to this hyperparameter and to alternative saliency-mask thresholds is discussed as a limitation.

  \section{Results}
\label{sec:results}

\subsection{Class Sensitivity Analysis}
\label{sec:results_sensitivity}

\par To evaluate whether existing saliency methods produce class-sensitive explanations, we compute the top-5\% feature agreement between saliency maps for the top-1 and top-2 predicted classes on the same input image. High agreement implies that the method highlights largely the same features regardless of the target class, violating the class-specificity criterion.

\par
\begin{figure*}[t]
    \centering
    \includegraphics[width=\textwidth]{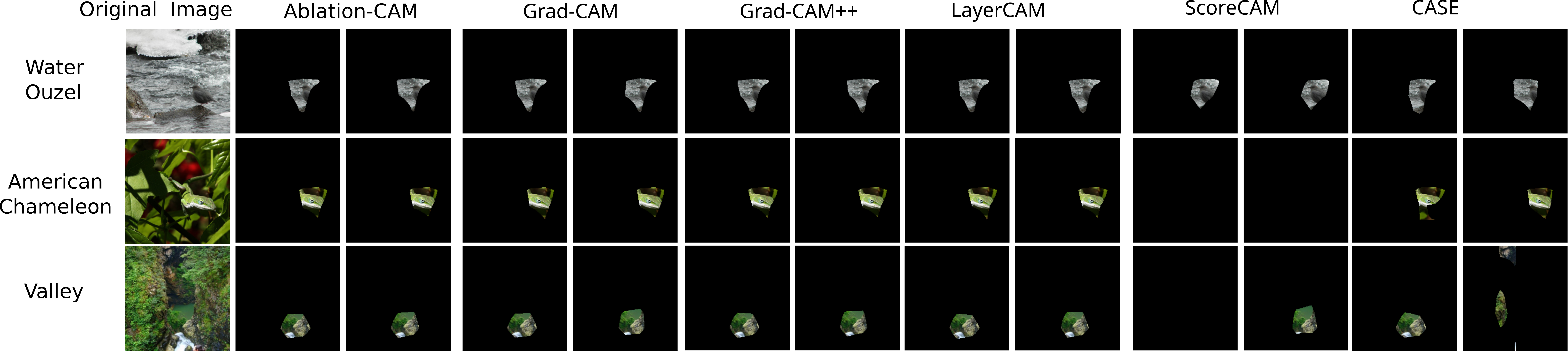}
    \caption{
        Class sensitivity comparison across saliency methods. 
        Each row shows the original image and saliency-based feature masking for the top 5 \% of the most salient pixels for the top-2 predicted classes.
        The methods compared are Grad-CAM, Grad-CAM++, ScoreCAM, AblationCAM, LayerCAM, and CASE. 
        For each method, the left masked output corresponds to the most probable class, and the right to the second-most probable class.
        The visible row headers identify the examples as \textit{Water ouzel}, \textit{American chameleon}, and \textit{Valley}, respectively; each row shows the top-1 and top-2 predicted-class explanations for that image. The class names in the row headers and image panels are the authoritative labels for the displayed examples.
        Methods like ScoreCAM often produce visually identical or absent explanations, highlighting a lack of class-specificity. 
        In contrast, CASE consistently isolates distinct evidence for each class label.
    }
    \label{fig:class_sensitivity_comparison}
\end{figure*}

 To complement the quantitative results, \autoref{fig:class_sensitivity_comparison} presents a visual comparison of masked saliency regions across methods. For each input, we show the masked top-1 and top-2 class explanations produced by Grad-CAM variants, ScoreCAM, LayerCAM, and CASE on a ResNet model. Methods like ScoreCAM and Grad-CAM produce visually indistinct explanations across labels, whereas CASE consistently isolates class-specific evidence.

\par We perform this analysis across two benchmark datasets (ImageNet and CIFAR-100) and four popular convolutional architectures (ResNet, DenseNet, VGG, ConvNeXt). For each method--model--dataset combination, we run a one-sided Wilcoxon signed-rank test to evaluate the null hypothesis that the median agreement is greater than or equal to 50\%. A statistically significant rejection of the null ($p < 0.05$) indicates that the method produces class-distinct explanations in that setting.

\par Table~\ref{tab:rq1-wilcoxon} reports the $p$-values. For each entry, if $p < 0.05$, we reject the null hypothesis and conclude that the method produces significantly different saliency maps for the top-1 and top-2 class labels. If $p \geq 0.05$, we fail to reject the null and cannot conclude that the explanations differ.

\begin{table}[t]
    \centering
    \caption{\label{tab:rq1-wilcoxon}
        Wilcoxon signed-rank test $p$-values for top-5\% agreement between saliency maps of top-1 and top-2 predicted class labels.
        Models are pretrained on ImageNet. Null hypothesis: median agreement $\geq 50\%$. 
        Bolded entries indicate significant rejection of the null at $p < 0.05$.
    }
    \begin{tabular}{l |c|c|c|c}
        \toprule
        \textbf{Method} & \textbf{ConvNeXt} & \textbf{DenseNet} & \textbf{ResNet} & \textbf{VGG} \\
        \midrule
        AblationCAM    & 1.       & $<.0001$ & 1.       & 1.       \\
        \texttt{CASE}    &   $<.0001$     &  $<.0001$ &  $<.0001$     &    $<.0001$    \\
        Grad-CAM       & 0.8734   & $<.0001$ & 1.       & 1.       \\
        Grad-CAM++     & 0.0179   & $<.0001$ & 1.       & $<.0001$ \\
        LayerCAM       & $<.0001$ & $<.0001$ & 1.       & 1.       \\
        ScoreCAM       & $<.0001$ & $<.0001$ & $<.0001$ & 1.       \\
        \bottomrule
    \end{tabular}
\end{table}

\par Across all architectures, \case consistently rejects the null hypothesis with high confidence ($p < 0.0001$), demonstrating robust class sensitivity. ScoreCAM also performs well on most models but fails on VGG. In contrast, traditional methods like Grad-CAM and AblationCAM frequently fail to distinguish between top-1 and top-2 saliency maps, especially on ResNet and VGG. Interestingly, DenseNet stands out as the most sensitive architecture: nearly all methods, even weak baselines, achieve significance. These trends support the claim that DenseNet’s internal representations are more class-discriminative, and highlight that \case provides the most consistent signal separation among the evaluated methods and CNN backbones.

\subsubsection{Robustness to Training Seeds (DenseNet on CIFAR-100)}
\label{sec:results_seed_variation}

\par To test whether the class sensitivity observed in DenseNet is robust to training randomness, we retrained DenseNet-201 on CIFAR-100 using five different random seeds. We then re-evaluated the saliency methods using the RQ1 procedure for each independently trained model.

\par
\begin{figure}[t]
    \centering
    \includegraphics[width=\columnwidth]{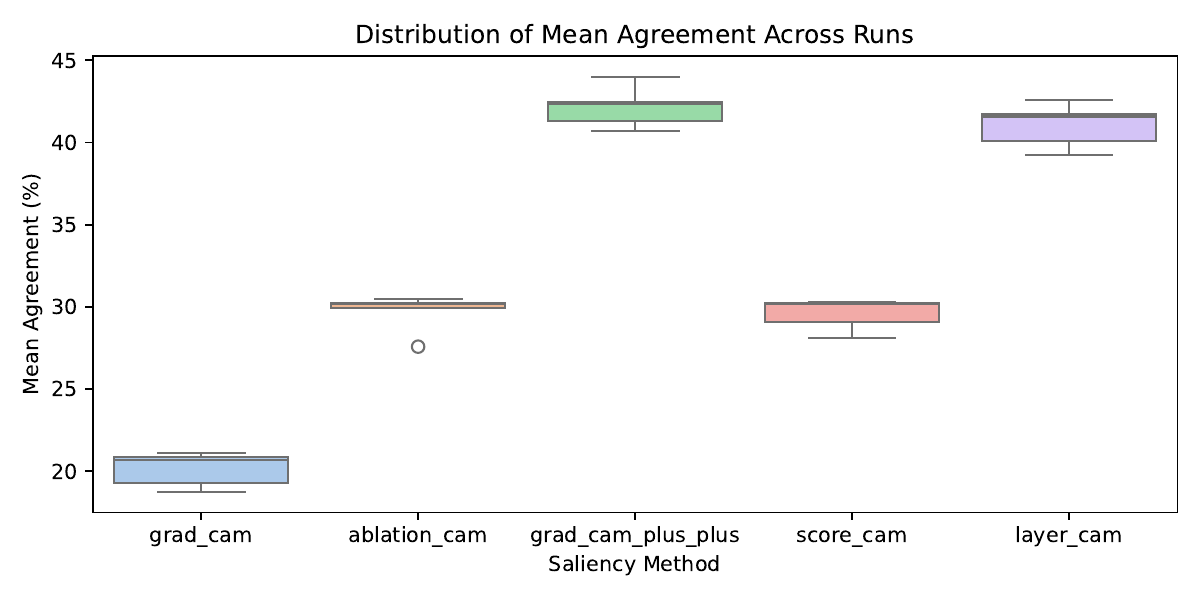}
    \caption{Mean agreement scores across seeds for each method (DenseNet, CIFAR-100). Lower values indicate more class-distinct saliency.}
    \label{fig:agreement_boxplot}
\end{figure}

 As shown in Figure~\ref{fig:agreement_boxplot}, the relative class sensitivity rankings of methods remain consistent across seeds, and all methods continue to reject the null hypothesis. These findings suggest that DenseNet’s ability to produce class-distinct saliency maps is a stable property of the architecture and not an artifact of optimization variability.

\subsection{Ablation Study: Channel Sparsity and Discriminative Structure}
\label{sec:results_channel_sparsity}

\par The results in RQ1 reveal that DenseNet exhibits uniquely consistent class sensitivity across saliency methods. To understand this behavior, we analyze the \textit{channel-wise activation patterns} of the layer used for saliency attribution in each architecture.

\par For each model, we extract activations from the target convolutional layer used in RQ1 and compute the number of channels that are active per input. A channel is considered active if its spatially pooled activation exceeds a small threshold $\tau$:

\begin{equation}
    \label{eq:activation}
    \hat{a}_n = \frac{1}{HW} \sum_{i=1}^H \sum_{j=1}^W A_n(i,j), \quad \text{Active if } \hat{a}_n > \tau,
\end{equation}

\noindent where $A_n$ denotes the activation map for channel $n$. For each class, we compute the \textit{mean number of active channels} across correctly predicted validation examples.

\begin{table}[t]
    \centering
    \caption{\label{tab:densenet-norm5}
        RQ1 class sensitivity results when using \texttt{features.norm5} (final normalization layer) instead of the final convolutional layer in DenseNet-201.
    }
    \begin{tabular}{l | c}
        \toprule
        \textbf{Method} & \texttt{DenseNet (norm5 layer)} \\
        \midrule
        AblationCAM    &  1. \\
        \texttt{CASE}   &  $<0.0001$ \\
        Grad-CAM       & 1. \\
        Grad-CAM++     & 1. \\
        LayerCAM       & 1. \\
        ScoreCAM       &  $<0.0001$ \\
        \bottomrule
    \end{tabular} 
\end{table}

\par
\begin{figure}[t!]
    \centering
    \includegraphics[width=\columnwidth]{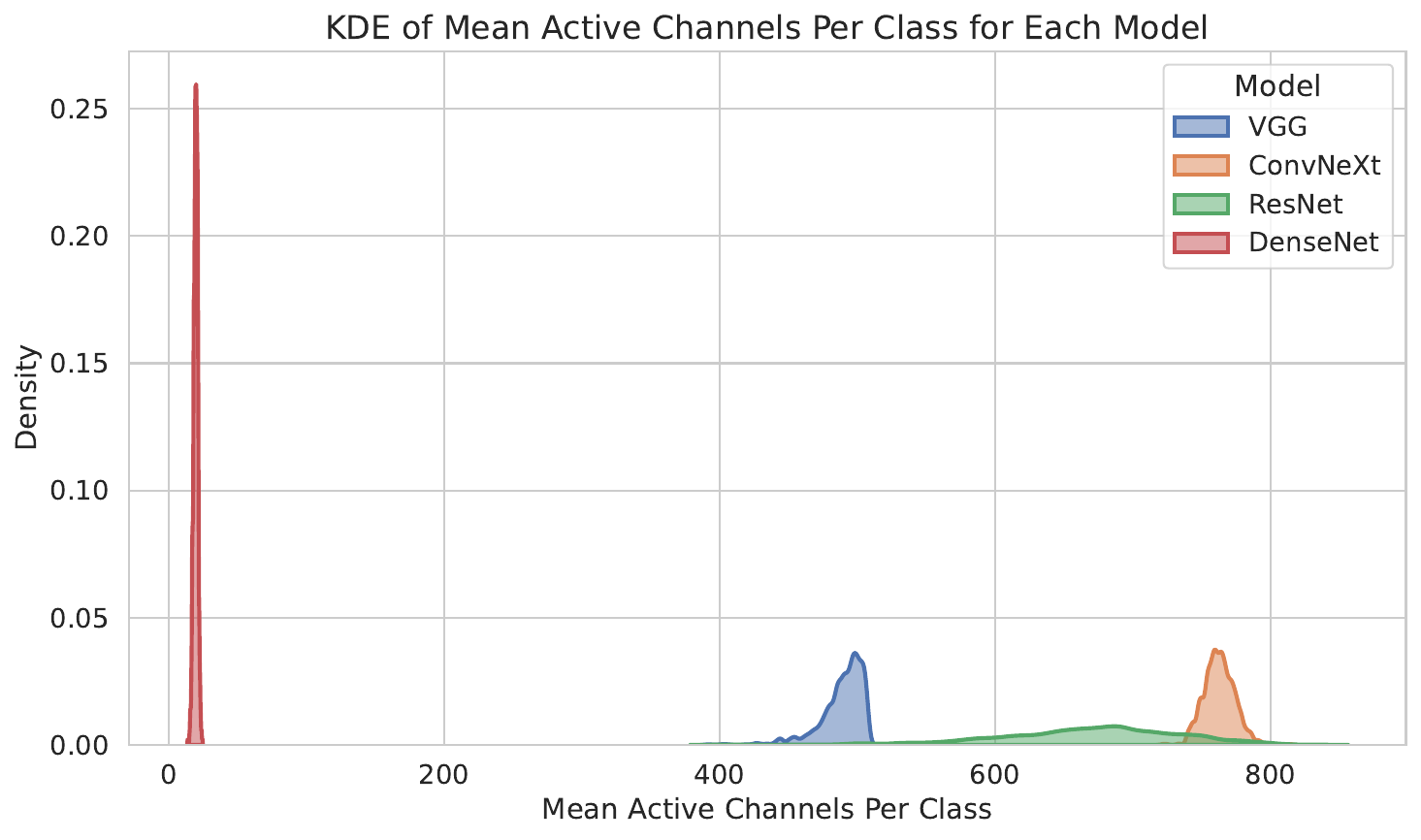}
    \caption{
        Distribution of per-class mean active channel counts across architectures.
        DenseNet shows the most sparse channel activations, with most classes activating fewer than 20 channels.
        In contrast, ResNet, VGG, and ConvNeXt activate substantially more channels on average.
    }
    \label{fig:channel_activity_kde}
    \vspace{-1em}
\end{figure}

 We observe a striking pattern in DenseNet. When using $\tau = 0.001$, the final convolutional layer (\texttt{features.denseblock4.denselayer32.conv2}) exhibits \textit{extremely sparse activation}, averaging fewer than 20 active channels per class. This is in stark contrast to other architectures, whose corresponding attribution layers activate several hundred channels per class on average. This trend is illustrated in \autoref{fig:channel_activity_kde}, which shows the kernel density estimates (KDE) of mean active channels per class across architectures.

\par
\begin{figure}[t!]
    \centering
    \includegraphics[width=\columnwidth]{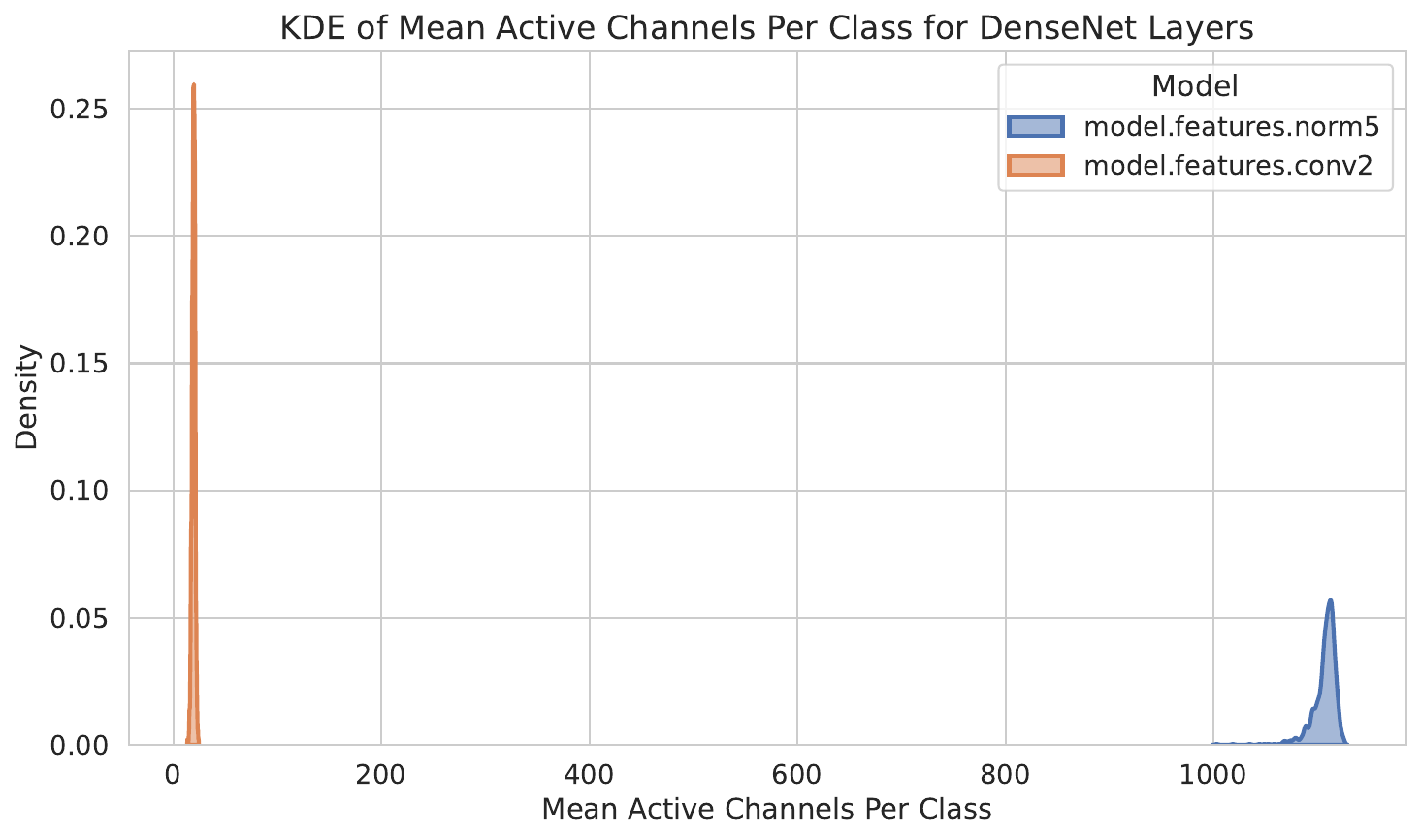}
    \caption{
        Kernel density estimate (KDE) of mean active channels per class for two attribution layers in DenseNet-201. 
        The final convolutional layer (\texttt{features.denseblock4.denselayer32.conv2}) exhibits extremely sparse activation, 
        with fewer than 20 active channels per class. In contrast, the normalization layer (\texttt{features.norm5}) activates nearly all channels. 
        Despite its sparsity, the final convolutional layer yields more class-sensitive saliency maps, underscoring the importance of 
        selecting structurally discriminative layers for attribution.
    }
    \label{fig:densenet_layers_kde}
\end{figure}

 To determine whether this sparsity was a general property of the DenseNet architecture or a consequence of layer selection, we repeated the analysis using an alternative target layer: \texttt{features.norm5}, the final batch normalization layer preceding global pooling. As shown in \autoref{fig:densenet_layers_kde}, this layer exhibits dramatically higher activation density—nearly all channels are active per class, with the KDE peaking around 1050 channels.
\par Notably, when we re-ran RQ1 using \texttt{features.norm5}, the final normalization layer in DenseNet-201, instead of the final convolutional layer, the results were markedly different. As shown in Table~\ref{tab:densenet-norm5}, most saliency methods failed to reject the null hypothesis, indicating that their explanations did not significantly differ across class labels. This stands in sharp contrast to the results reported in Table~\ref{tab:rq1-wilcoxon} for the convolutional layer, where nearly all methods achieved significant class sensitivity. These findings reinforce the conclusion that saliency quality is highly sensitive to layer choice, and that DenseNet’s class-distinct explanations are driven by the sparse and selective representations encoded in its final convolutional layer. Notably, \case retains its class sensitivity even under this more diffuse, high-activation regime.

\par These findings suggest that DenseNet’s final convolutional layer encodes \textit{highly selective and class-discriminative features}. While fewer channels activate, the \textit{identity} of those channels varies meaningfully with class, enabling saliency methods to produce distinct explanations. This aligns with the DenseNet architecture’s design: its dense connectivity may lead to late-stage layers specializing in compact, high-value features with minimal redundancy.

\par In contrast, architectures like VGG and ConvNeXt exhibit broad activation across many channels, likely reflecting \textit{shared or distributed features} that dilute class-specific signals in the saliency maps. Thus, we posit that \textit{channel-level sparsity coupled with discriminative selectivity} explains why DenseNet supports stronger class-distinct explanations, even when using a layer with minimal activation, whereas dense activation may obscure discriminative attribution.

\textbf{}

\subsection{Explanation Fidelity}
\label{sec:results_fidelity}

\par To assess the faithfulness of saliency methods, we measure the change in classification confidence after ablating the top-$k$ most salient pixels identified by each method. For each image, we compute the softmax confidence of the predicted class before and after ablation. The difference reflects the influence of the identified regions on the model's output.

\par For each method, we compute the average confidence drop and compare it to that of \case using a paired t-test. The null hypothesis states that the mean confidence drop of the method is greater than or equal to that of \case. A statistically significant $p$-value ($p < 0.05$) supports rejecting the null in favor of the alternative, that \case yields a higher mean drop.

\par \autoref{tab:rq2-confidence-drop} presents the results across all four architectures. On ConvNeXt, \case significantly outperforms AblationCAM and LayerCAM, while showing no statistical difference from Grad-CAM, Grad-CAM++, or ScoreCAM. On DenseNet and ResNet, no baseline method differs significantly from \case, indicating comparable levels of confidence drop. On VGG, three methods, Grad-CAM++, LayerCAM, and ScoreCAM, yield significantly higher drops than \case, while the remaining methods are not statistically distinguishable. We treat these VGG results as concrete failure cases for \case under the perturbation-based fidelity metric, and analyze this tradeoff in Section~\ref{sec:discussion-rq2}.

\begin{table*}[t]
    \centering
    \caption{
        Mean confidence drop and standard deviation after ablating the top-$k$ salient pixels for each method on four architectures. Higher drop implies greater attribution faithfulness. $p$-values are from paired t-tests comparing each method to \texttt{CASE} (per model); bolded values indicate the highest mean confidence drop for a given model.
    }
    \label{tab:rq2-confidence-drop}
    \vspace{0.5em}
    \resizebox{\textwidth}{!}{%
    \begin{tabular}{l|ccc|ccc|ccc|ccc}
        \toprule
        \multirow{2}{*}{\textbf{Method}} &
        \multicolumn{3}{c|}{\textbf{ConvNeXt}} &
        \multicolumn{3}{c|}{\textbf{DenseNet}} &
        \multicolumn{3}{c|}{\textbf{ResNet}} &
        \multicolumn{3}{c}{\textbf{VGG}} \\
        & \textbf{Mean} & \textbf{SD} & \textbf{$p$} 
        & \textbf{Mean} & \textbf{SD} & \textbf{$p$}
        & \textbf{Mean} & \textbf{SD} & \textbf{$p$}
        & \textbf{Mean} & \textbf{SD} & \textbf{$p$} \\
        \midrule
        AblationCAM    & .1696 & .2409 & .00568  & .1140 & .2272 & .65943 & .1339 & .2532 & .21707 & .1354 & .2473 & .51132 \\
        \textbf{CASE}  & .2052 & .3042 & --      & .1163 & .2274 & --     & $\textbf{.1411}$ & .2561 & --     & .1317 & .2420 & -- \\
        Grad-CAM       & .1919 & .2758 & .3544   & .1131 & .2226 & .53783 & .136  & .2501 & .37022 & .1328 & .2433 & .84450 \\
        Grad-CAM++     & .1872 & .2860 & .22602  & .1244 & .2417 & .13240 & .1395 & .2584 & .78072 & .1020 & .2096 & $<.0001$ \\
        LayerCAM       & .1686 & .2783 & .00988  & \textbf{.1248} & .2393 & .10838 & .1405 & .2603 & .92128 & \textbf{.1539} & .2650 & $<.0001$ \\
        ScoreCAM       & \textbf{.2059} & .3104 & .9665   & \textbf{.1248} & .2366 & .10598 & .1349 & .2536 & .28969 & .1473 & .2590 & .00645 \\
        \bottomrule
    \end{tabular}
    }
\end{table*}

  \section{Discussion}
\label{sec:discussion}

\subsection{RQ1: Class Sensitivity of Saliency Methods}
\label{sec:discussion-rq1}

\par The results in \autoref{tab:rq1-wilcoxon} demonstrate that the ability of saliency methods to produce class-specific explanations varies substantially across model architectures. DenseNet stands out for its consistent rejection of the null hypothesis across all tested saliency methods, indicating robust class-distinct attribution. In contrast, architectures like VGG and ResNet fail to produce class-sensitive maps under several methods, notably Grad-CAM and AblationCAM. ScoreCAM achieves broader consistency, performing well on both DenseNet and ResNet, while methods such as LayerCAM and Grad-CAM++ show stronger dependency on the underlying architecture.

\par These trends suggest that class sensitivity is shaped not only by the saliency algorithm but also by architectural factors. DenseNet's dense connectivity and feature reuse mechanisms may enable more granular and disentangled activations that preserve class-conditional variation late into the network. However, isolating which architectural features are responsible required targeted ablations.

\subsubsection*{Ablation: Robustness to Training Seed Variation}
\par To test whether the class sensitivity of DenseNet might be a byproduct of training stochasticity, we retrained DenseNet-201 on CIFAR-100 using five random seeds. As shown in \autoref{fig:agreement_boxplot}, the mean agreement distributions across methods remain consistent across all runs. All saliency methods continue to reject the null hypothesis under the RQ1 test, confirming that the class sensitivity is not merely due to favorable initialization but is instead a robust property of the model. This affirms the role of architecture in enabling class-discriminative explanations.

\subsubsection*{Ablation: The Role of Layer Selection}
\par We further investigated whether DenseNet’s class sensitivity could be attributed to the specific layer used for attribution. While conventional practice often selects the final convolutional layer for saliency computation, we found that \texttt{features.denseblock4.denselayer32.conv2}, the final convolution in DenseNet-201, exhibited unusually sparse activation. As shown in \autoref{fig:channel_activity_kde}, this layer activates, on average, fewer than 20 channels per class when using a low threshold of $\tau=0.001$, compared to several hundred in VGG, ResNet, and ConvNeXt.

\par Despite this sparsity, RQ1 results reveal that saliency methods still produce highly class-distinct explanations when using this layer. We interpret this to mean that although few channels are active, their activation patterns are highly class-specific. In other words, DenseNet leverages a compact, selective set of features for each class, which aids saliency methods in isolating discriminative regions.

\par Importantly, when we re-ran RQ1 using an alternative attribution layer, \texttt{features.norm5}, the results were less consistent. Not all methods were able to reject the null hypothesis in this configuration. This shift confirms that layer choice significantly affects explanation quality, even within the same architecture.

\par While \texttt{features.norm5} may appear more appropriate due to its proximity to the classifier and its aggregation of upstream features, our findings reveal that this aggregation may inadvertently homogenize activations across classes. As shown in \autoref{fig:densenet_layers_kde}, \texttt{norm5} activates nearly all channels uniformly, reducing the saliency method’s ability to isolate class-specific signals. In contrast, the final convolutional layer, though sparsely activated, preserves more localized and class-distinct activation patterns. This highlights a key insight: layers closer to the classifier are not always optimal for attribution, especially when their role is to smooth or normalize a wide range of features. Selective, structurally discriminative layers may offer greater fidelity in class-sensitive explanation tasks.

\par The ablation in ~\autoref{tab:densenet-norm5} underscores that effective attribution depends not only on proximity to the classifier, but also on the structural selectivity of the chosen layer. These results motivate the need to treat attribution layer selection as a principled design decision, rather than a fixed convention.

\subsubsection*{Analysis: Consistency of \case Across Architectures}

\par Notably, \case is the only evaluated method that consistently rejects the null hypothesis across all four model architectures in RQ1. This consistency is especially striking on VGG and ResNet, where most traditional methods fail to distinguish between class labels. While ScoreCAM and Grad-CAM++ perform well on specific models, their sensitivity is not uniformly observed. In contrast, \case provides a robust mechanism for isolating class-specific evidence within the CNN settings studied here.

\par This consistent class sensitivity suggests that the contrastive subtraction in \case effectively suppresses overlapping or shared attribution components, allowing more discriminative features to emerge, even in architectures where representations are otherwise more entangled. Specifically, \case operates by isolating a refined subset of the discriminative feature set $ D_u$ (~\autoref{eq:discriminative}) by subtracting the shared components identified in contrast classes $ \mathcal{V} $ (~\autoref{eq:contrast_class}). This process approximates the uniquely discriminative feature set $ U_u^{\mathcal{V}} $ (~\autoref{eq:udf}), which excludes any features that also support confused alternatives. The effectiveness of this subtraction is evidenced by \case’s consistent rejection of the null in RQ1, even on architectures where other methods fail. Importantly, this performance is achieved without architecture-specific tuning across the tested CNN backbones.

\par Together, these ablations support the following main findings:
\begin{itemize}
    \item \textbf{Layer selection is critical:} The choice of attribution layer strongly determines whether saliency maps are class-specific. Even within the same model, different layers can produce radically different results under RQ1.
    \item \textbf{Channel activity analysis is diagnostic:} Measuring mean active channels per class reveals which layers exhibit class-variant structure. Sparse but selective layers, like DenseNet’s final convolution, are particularly effective for producing distinct saliency maps.
    \item \textbf{\case yields robust class sensitivity:} Unlike other methods, \case consistently produces class-distinct explanations across architectures, underscoring the effectiveness of its contrastive formulation.
\end{itemize}

\subsection{RQ2: Explanation Fidelity of Saliency Methods}
\label{sec:discussion-rq2}

\par The results in \autoref{tab:rq2-confidence-drop} indicate that \case achieves statistically significant confidence drops compared to AblationCAM and LayerCAM on ConvNeXt, and performs comparably to the strongest baselines on DenseNet and ResNet. On VGG, Grad-CAM++, LayerCAM, and ScoreCAM achieve higher confidence drops than \case, and all three differences are statistically significant.

\par These findings suggest that \case performs on par with the most widely used attribution methods overall, while providing substantial improvements over weaker baselines. However, attribution fidelity, as measured by the drop in confidence when salient regions are masked, does not always align with class sensitivity. A method may highlight features that are distinct across classes (indicating high RQ1 sensitivity), yet some of these features may exert less influence on the model’s output confidence than shared, highly activated regions. As a result, saliency maps that emphasize uniquely discriminative features may yield relatively modest confidence drops, even if they are more class-specific and interpretable.

\par Conversely, saliency methods that emphasize broadly activated or background regions may achieve larger confidence drops when those areas are occluded, despite failing to differentiate between classes. This illustrates a key tension: optimizing for overall model confidence may sometimes come at the cost of specificity, and vice versa.

\par The combined results from RQ1 and RQ2 suggest that \case addresses a major limitation of existing methods, class-insensitivity, without sacrificing attribution fidelity. While it does not uniformly outperform other methods in confidence drop, its consistently high class specificity represents a complementary strength. This tradeoff makes \case a valuable addition to the saliency toolbox, particularly in applications that demand clear differentiation between competing class explanations.

\subsubsection*{Failure Cases and Tradeoffs}
\par The VGG results provide the clearest failure case for \case under the perturbation-based fidelity metric: Grad-CAM++, LayerCAM, and ScoreCAM produce larger confidence drops than \case, with the former two differences statistically significant. This does not contradict the class-sensitivity result; rather, it indicates that uniquely discriminative regions are not always the regions whose removal maximally reduces confidence. In models where class evidence is diffuse or shared across competing labels, perturbing common high-activation regions can cause a larger confidence drop, even when those regions are less informative for distinguishing between classes.

\par These cases clarify the intended use of \case. It is most beneficial when the explanatory goal is to identify what separates one plausible class hypothesis from another. When the goal is instead to maximize confidence drop under a fixed perturbation budget, non-contrastive methods may be preferable for some architectures.

\begin{figure*}[h!]
\centering

\begin{subfigure}{0.48\linewidth}
    \centering
    \includegraphics[width=\linewidth]{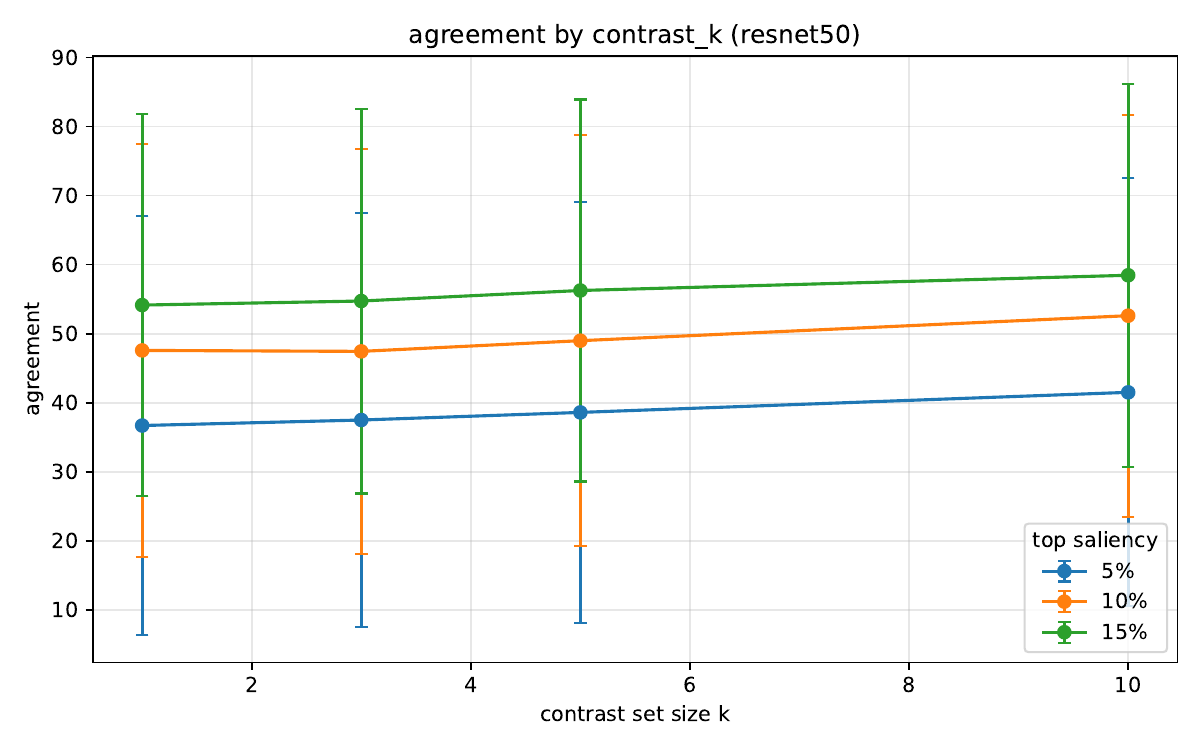}
    \caption{ResNet-50: Agreement}
\end{subfigure}
\hfill
\begin{subfigure}{0.48\linewidth}
    \centering
    \includegraphics[width=\linewidth]{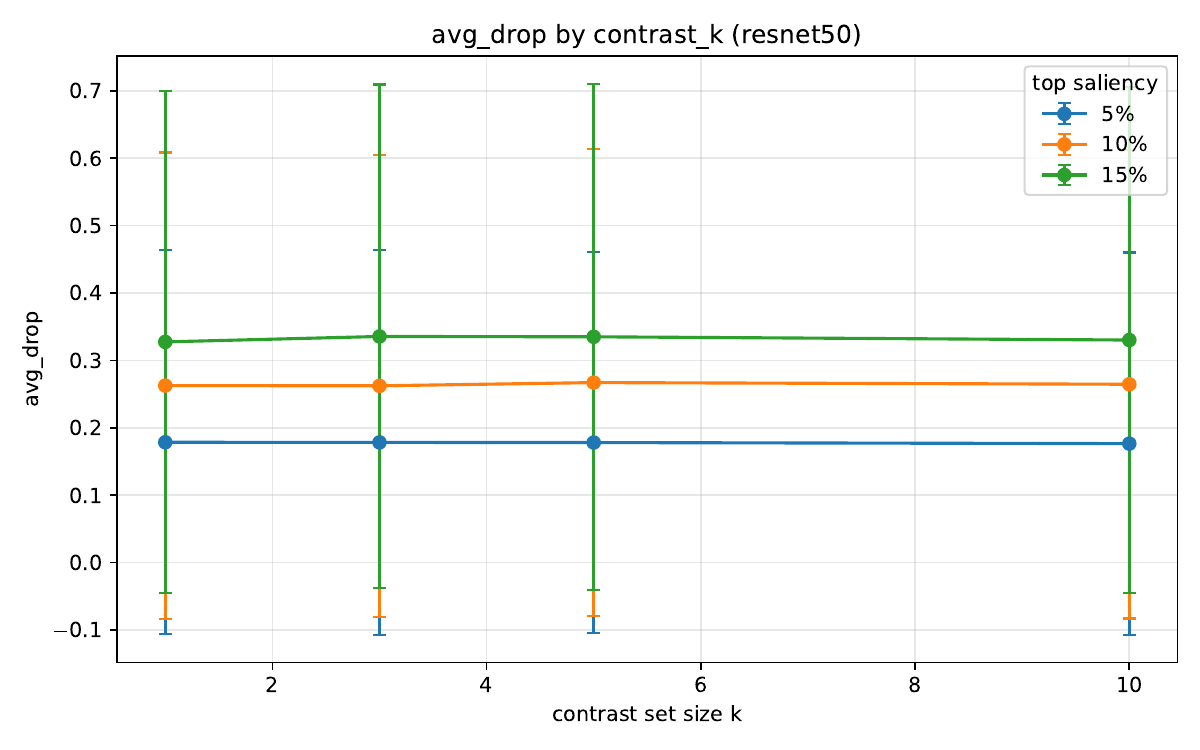}
    \caption{ResNet-50: Confidence Drop}
\end{subfigure}

\vspace{0.5em}

\begin{subfigure}{0.48\linewidth}
    \centering
    \includegraphics[width=\linewidth]{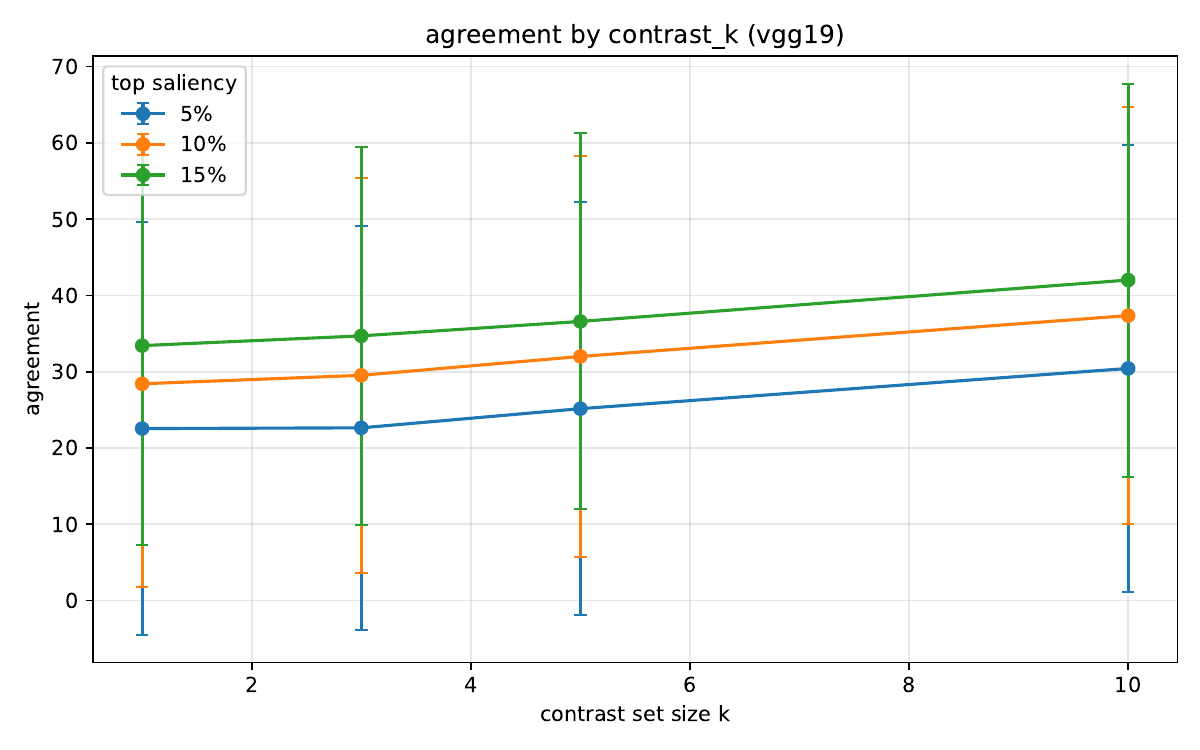}
    \caption{VGG-19: Agreement}
\end{subfigure}
\hfill
\begin{subfigure}{0.48\linewidth}
    \centering
    \includegraphics[width=\linewidth]{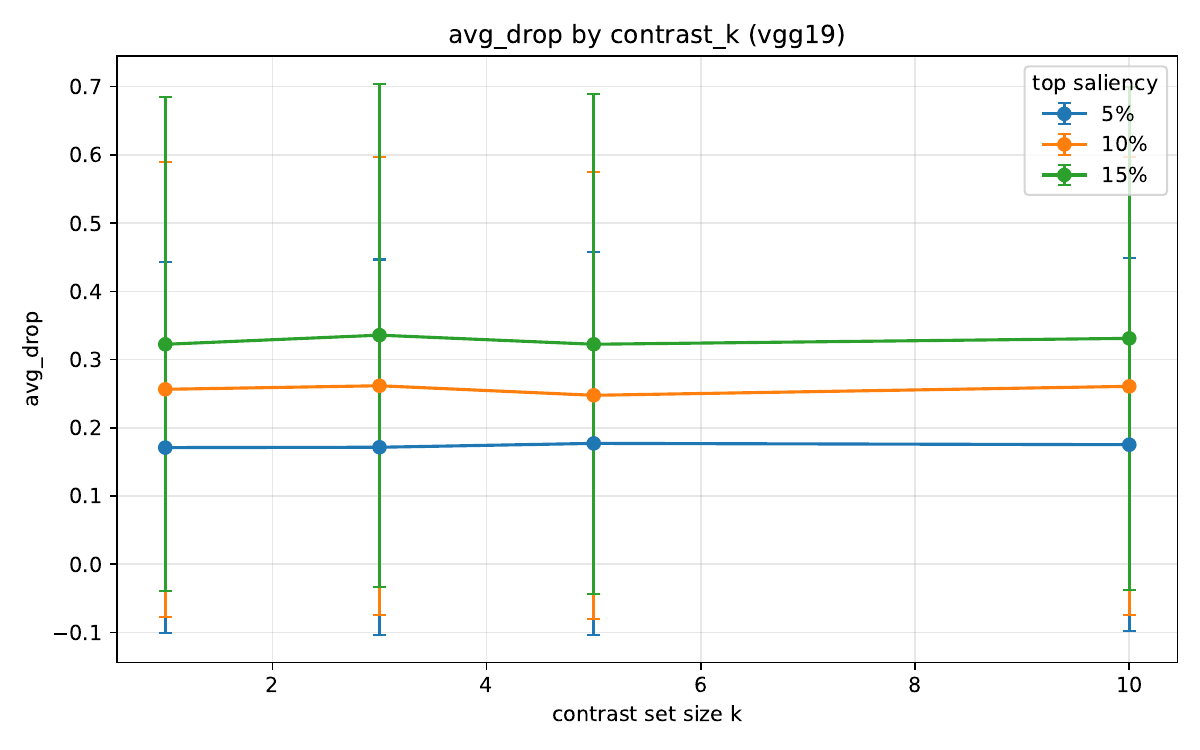}
    \caption{VGG-19: Confidence Drop}
\end{subfigure}

\vspace{0.75em}

\begin{subfigure}{0.48\linewidth}
    \centering
    \includegraphics[width=\linewidth]{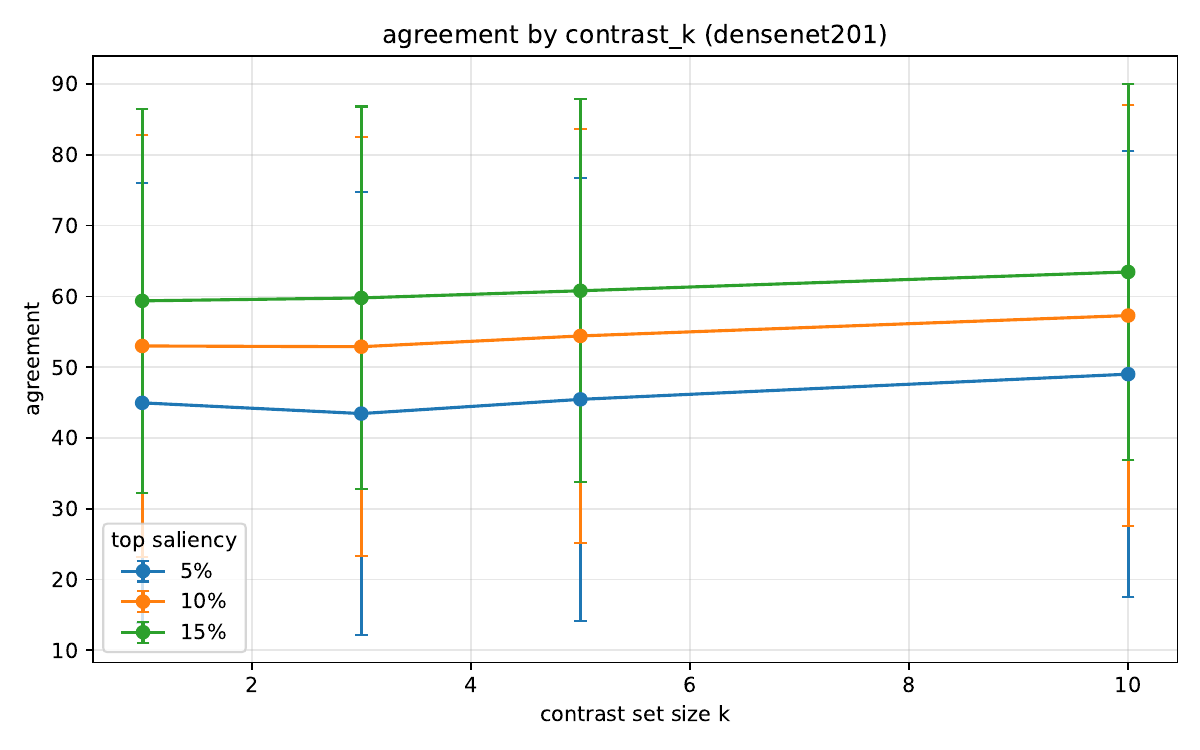}
    \label{fig:case_ablation_k_densenet_agreement}
    \caption{DenseNet-201: Agreement}
\end{subfigure}
\hfill
\begin{subfigure}{0.48\linewidth}
    \centering
    \includegraphics[width=\linewidth]{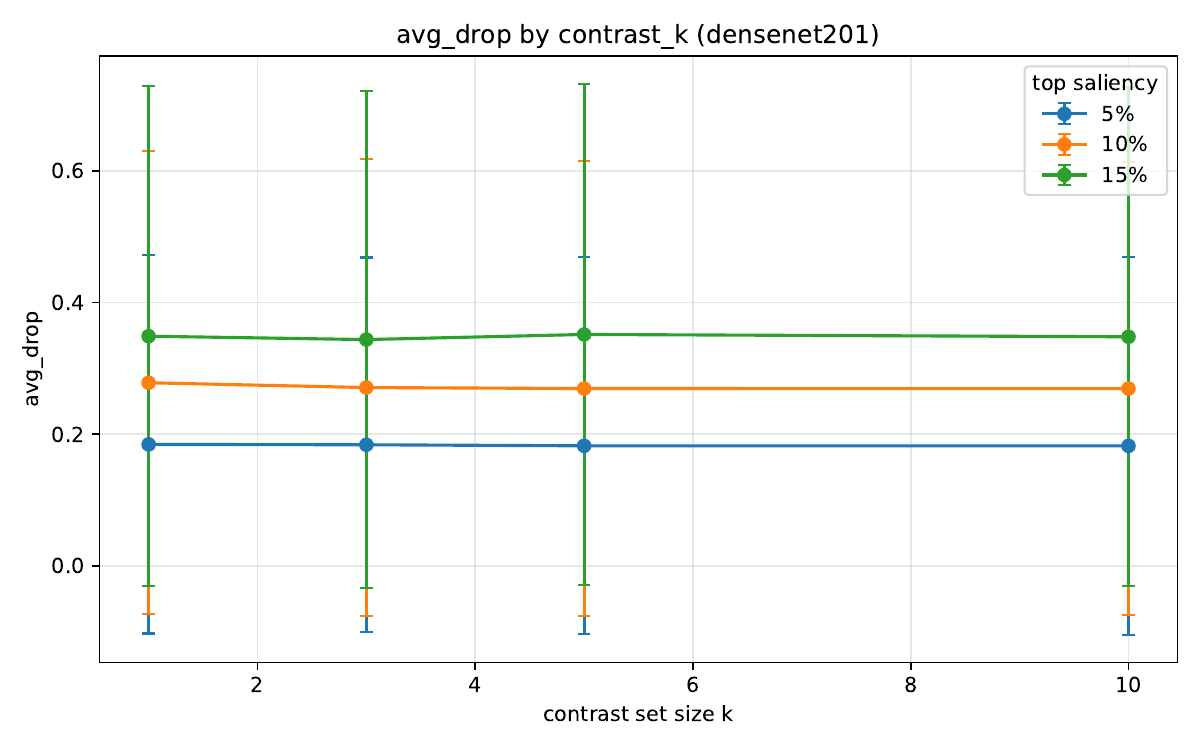}
    \label{fig:case_ablation_k_densenet_drop}
    \caption{DenseNet-201: Confidence Drop}
\end{subfigure}

\vspace{0.5em}

\begin{subfigure}{0.48\linewidth}
    \centering
    \includegraphics[width=\linewidth]{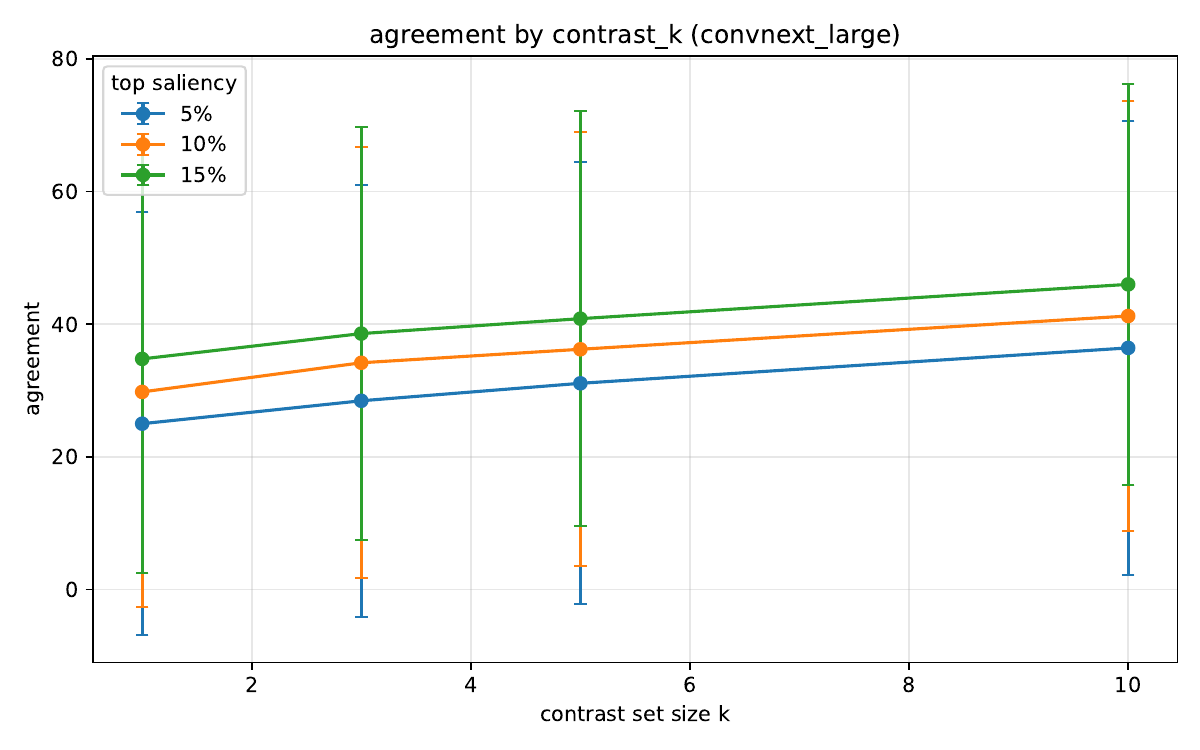}
    \caption{ConvNeXt: Agreement}
\end{subfigure}
\hfill
\begin{subfigure}{0.48\linewidth}
    \centering
    \includegraphics[width=\linewidth]{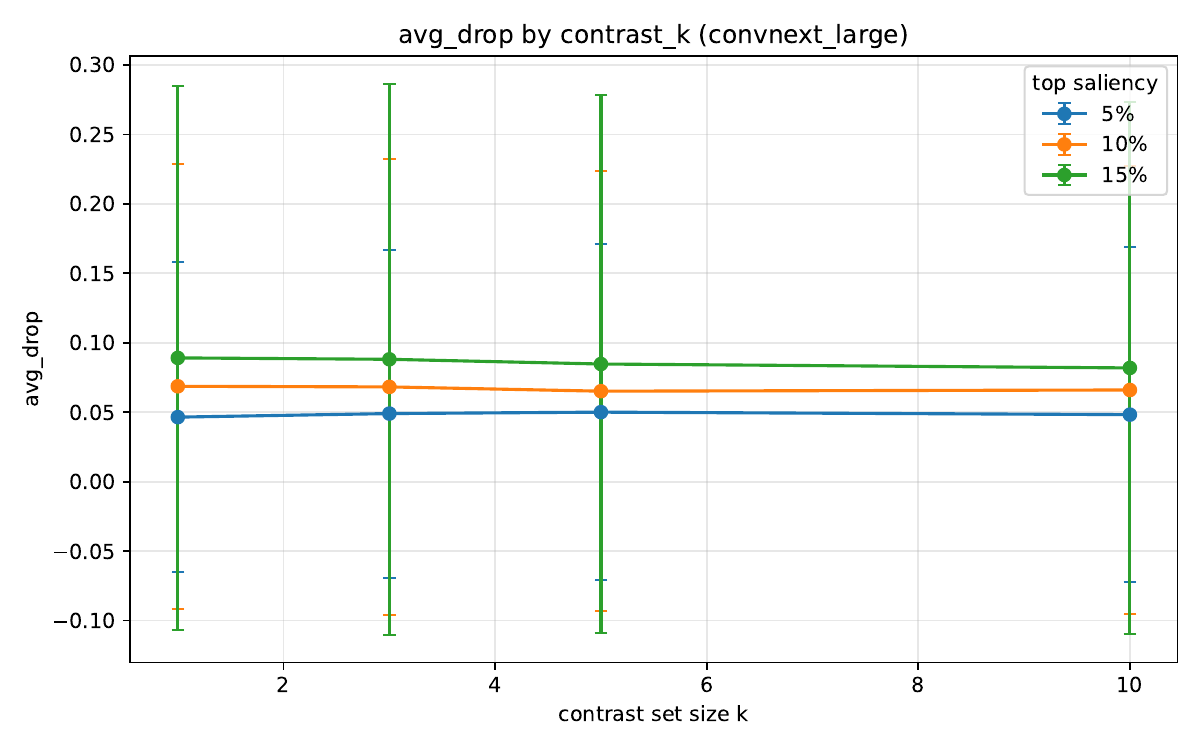}
    \caption{ConvNeXt: Confidence Drop}
\end{subfigure}

\caption{
Ablation of contrast set size $k$ and saliency mask threshold $m$ across architectures. 
Left plots show class-sensitivity (agreement; lower is better), and right plots show perturbation fidelity (confidence drop; higher is better).
}
\label{fig:case_ablation_k}
\end{figure*}

\subsection*{Ablation Study: Effect of Contrast Set Size $k$ and Mask Threshold $m$}
\label{sec:results_ablation_topk}

We evaluate the sensitivity of \case to the size of the contrast set by varying $k$, the number of confused classes used to construct the contrast signal, with k $\in \{1, 3, 5, 10\}$. This ablation tests whether \case depends critically on the number of contrast classes or remains stable across different contrast-set sizes.

For each model and input image, we generate \case saliency maps for the top-1 and top-2 predicted classes. We measure class sensitivity using the feature agreement metric defined in~\autoref{eq:feature_agreement}, which computes the overlap between the top-$m$\% most salient pixels of the two maps. Lower agreement indicates stronger class-specificity.

We evaluate perturbation-based fidelity by ablating the top-$m$\% salient pixels identified for the top-1 prediction and measuring the resulting drop in predicted-class confidence, as defined in~\autoref{eq:rq_two_null_hypothesis}.

To assess robustness to the evaluation budget, we vary the saliency mask size $m \in \{ 5\%, 10\%, 15\% \}$, where $m$ controls the proportion of highest-saliency pixels retained or ablated. For each configuration $(k, m)$, agreement and confidence drop are computed on a per-example basis and aggregated over the evaluation set.

This ablation allows us to examine whether increasing the number of contrast classes strengthens the contrastive signal, or whether small contrast sets are sufficient to isolate discriminative features.

\subsubsection[Effect of Contrast Set Size Across Models]{Effect of Contrast Set Size Across Models}
\label{sec:results_ablation_topk_results}

\autoref{fig:case_ablation_k} summarizes the effect of varying the contrast set size $k$ and mask size $m$ across all architectures. Across models, agreement (left column) consistently increases with $m$, reflecting the expected increase in overlap when larger portions of the saliency maps are considered. In contrast, agreement shows little systematic dependence on $k$: across $k \in \{1,3,5,10\}$, the curves remain broadly similar, with only minor and inconsistent differences (e.g., slight separation for VGG-19) in~\autoref{fig:case_ablation_k}. This suggests that CASE obtains most of its contrastive benefit from a small number of confused classes, and that increasing the contrast set size does not reliably improve class-specificity.

A similar pattern holds for perturbation-based fidelity (right column). Confidence drop increases with $m$, as ablating larger salient regions produces a greater effect on model predictions. In contrast, varying $k$ has negligible impact, with nearly overlapping curves across all architectures (e.g., DenseNet-201 and ConvNeXt in~\autoref{fig:case_ablation_k}).

Overall, these results suggest that \case is not strongly sensitive to the contrast set size within the tested range. Increasing $k$ does not yield consistent improvements in class-specificity or perturbation-based fidelity, suggesting that small contrast sets are sufficient to capture most of the relevant contrastive signal. Instead, the largest and most predictable source of variation is the mask size $m$, which directly controls the attribution budget used for agreement and ablation.

  \section{Conclusion and Future Directions}
\label{sec:conclusion}

\par We presented \case, a contrastive activation-based saliency method that isolates class-specific signals by suppressing attribution shared with competing labels. Unlike conventional CAM-style techniques, \case  consistently generates visual explanations that are more class-distinct and better separate competing class evidence.

\par To validate this, we introduced a diagnostic framework for measuring class sensitivity via top-$k$ feature agreement, and assessed explanation fidelity through confidence drop analysis. Across the evaluated CNN models and datasets, \case provides the most consistent class specificity and generally competitive attribution faithfulness.

\par Our findings also highlight the overlooked influence of architectural features, such as activation sparsity and layer selectivity, on the quality of visual explanations. These insights suggest that saliency methods and model design should be considered jointly, not in isolation.

\par Future work will focus on adaptive strategies for selecting contrast classes, optimizing runtime, and extending \case to more complex tasks like multi-label classification and multimodal reasoning. Ultimately, we see \case as a step toward saliency methods that not only visualize what a model sees, but explain what truly sets its decisions apart.

  \section{Limitations}
\label{sec:limitations}

\subsection*{Formulation Constraints}
\par While \case improves class sensitivity and maintains competitive attribution fidelity in our experiments, several limitations remain. First, the method relies on selecting contrastive classes using a confusion matrix, which assumes stable and non-overlapping class relationships. This assumption may break down in low-data regimes or highly imbalanced classification settings.

\subsection*{Evaluation Constraints}
\par Our evaluation focuses on single-label natural-image classification with convolutional architectures because \case operates within the CAM framework, which relies on spatial convolutional feature maps for attribution. Transformer-based architectures represent images using tokenized attention representations rather than convolutional feature maps, and therefore require different attribution mechanisms. Similarly, extending \case to multi-label or multimodal settings would require redefining the contrast formulation when multiple target classes or modalities are present. We therefore focus on the CNN single-label setting where CAM-based explanations are most commonly applied, and leave these extensions to future work. As a result, our claims should be read as evidence for the CNN image-classification setting rather than as a general statement about all vision or multimodal models.

\par The empirical baseline set is also limited to widely used activation-based CAM methods. This scope is intentional: \case is designed as a CAM-style method, so Grad-CAM, Grad-CAM++, ScoreCAM, AblationCAM, and LayerCAM provide the most direct comparisons within the same activation-based attribution framework. Related contrastive explanation methods such as Contrastive LRP and CWOX are discussed conceptually, but they are not included as quantitative baselines because they operate under different explanation paradigms: C-LRP is tied to the LRP relevance-propagation framework, while CWOX uses a more general contrastive explanation mechanism rather than a CAM-style activation-map formulation. This limits the strength of any claim that \case is broadly superior to all contrastive explanation approaches. A more complete comparison would require implementations of these methods under the same model, dataset, attribution-layer, and perturbation protocols.

\subsubsection*{Computational Overhead}
\par While \case introduces minimal architectural dependencies, it incurs additional computational cost relative to standard CAM methods. This is primarily due to the need to compute gradients with respect to multiple contrast classes and perform vector projections at inference time. The added cost scales with the number of contrastive classes $k$; our main experiments use $k = 3$ , but in~\autoref{sec:results_ablation_topk} we sweep through $k \in \{ 1,3,5, 10 \}$. Optimizing the gradient computation pipeline and exploring approximation strategies (e.g., gradient caching or subspace projection) are potential directions for reducing runtime overhead in future deployments.


  \clearpage
  \bibliographystyle{plainnat}
  \bibliography{main}
  \clearpage
\onecolumn
\appendix
\section{Appendix}

\subsection{Agreement Score Distributions by Method}
\label{sec:appendix_agreement_dist}

To complement the statistical analysis reported in Section~\ref{sec:results_sensitivity}, we visualize the distribution of class agreement scores for each saliency method across all tested architectures. These histograms represent the percentage overlap between top-1 and top-2 saliency maps, aggregated over all validation images.

Each plot shows the relative frequency of agreement values (in 5\% bins), with a red dashed line marking the 50\% threshold used in our Wilcoxon signed-rank test. Distributions heavily concentrated below this line indicate strong class sensitivity, whereas those concentrated above suggest class-insensitive attribution behavior.

We present these distributions for each of the four architectures evaluated in the main paper: VGG, ResNet, ConvNeXt, and DenseNet. For DenseNet, we include two variants: one using the final convolutional layer (\texttt{conv2}), and another using the final normalization layer (\texttt{norm5}). This dual analysis allows us to visualize how attribution quality shifts with layer selection, supporting our ablation findings in Section~\ref{sec:results_channel_sparsity}.

\begin{figure*}[!ht]
    \centering
    \textbf{Feature Agreement Score Distributions for VGG}\\[1em]
    \begin{minipage}[b]{0.30\textwidth}
        \centering
        \includegraphics[width=\linewidth]{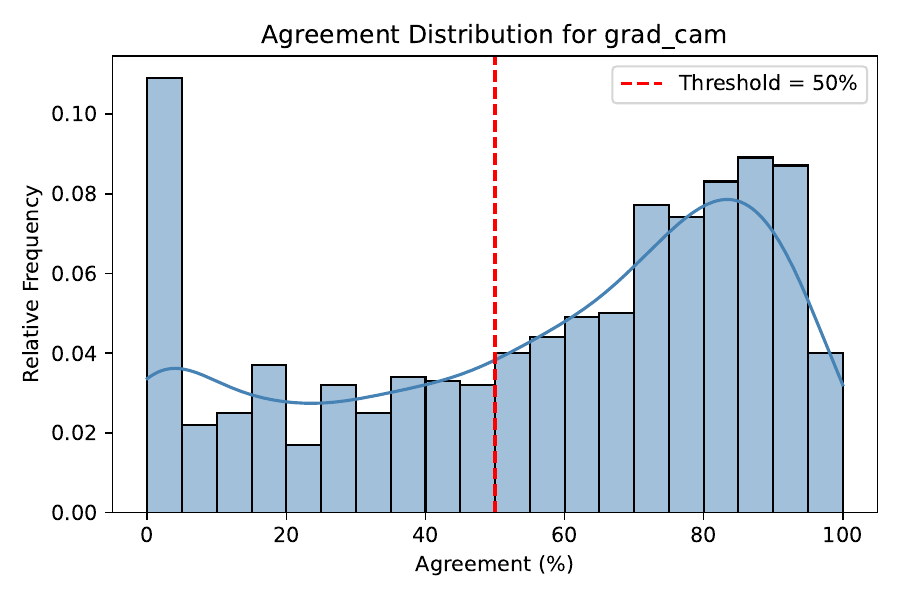}
        \\ \texttt{Grad-CAM}
    \end{minipage}
    \hfill
    \begin{minipage}[b]{0.30\textwidth}
        \centering
        \includegraphics[width=\linewidth]{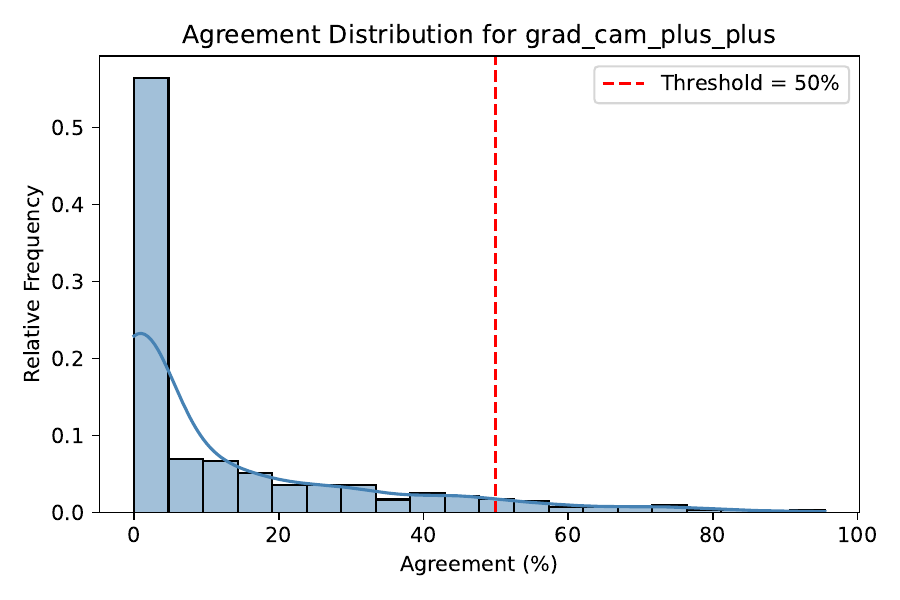}
        \\ \texttt{Grad-CAM++}
    \end{minipage}
    \hfill
    \begin{minipage}[b]{0.30\textwidth}
        \centering
        \includegraphics[width=\linewidth]{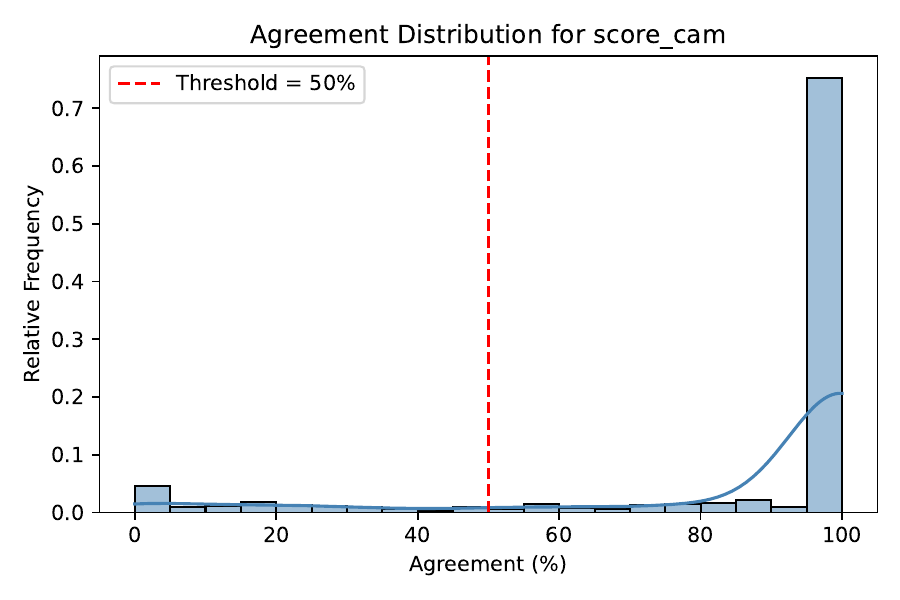}
        \\ \texttt{ScoreCAM}
    \end{minipage}

    \vspace{1em}

    \begin{minipage}[b]{0.30\textwidth}
        \centering
        \includegraphics[width=\linewidth]{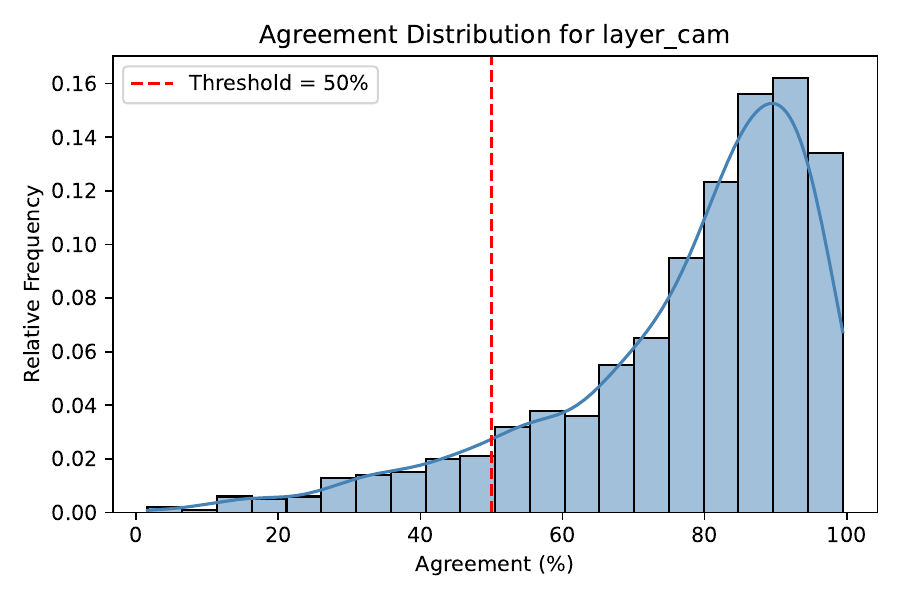}
        \\ \texttt{LayerCAM}
    \end{minipage}
    \hfill
    \begin{minipage}[b]{0.30\textwidth}
        \centering
        \includegraphics[width=\linewidth]{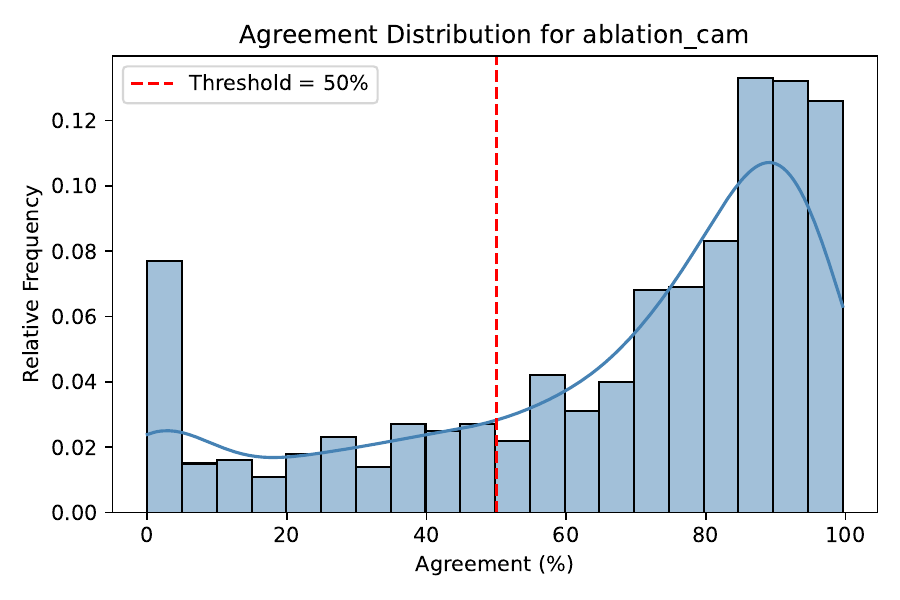}
        \\ \texttt{AblationCAM}
    \end{minipage}
    \hfill
    \begin{minipage}[b]{0.30\textwidth}
        \centering
        \includegraphics[width=\linewidth]{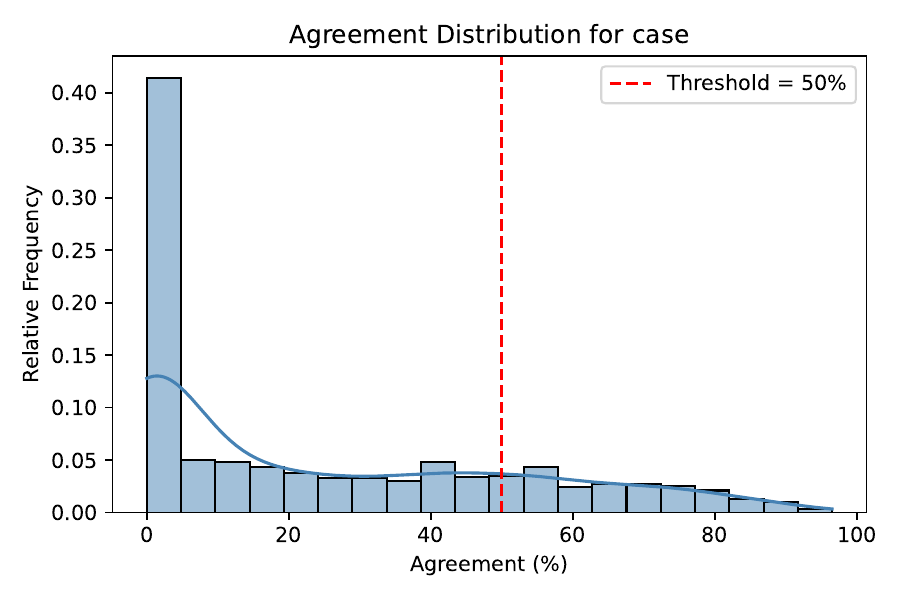}
        \\ \texttt{CASE}
    \end{minipage}

    \caption{
        Agreement score distributions for all saliency methods (VGG). Each histogram shows the distribution of top-5\% feature overlap between saliency maps generated for top-1 and top-2 class labels. The red dashed line denotes the 50\% threshold used in our Wilcoxon test for class sensitivity.
    }
    \label{fig:vgg_agreement_distributions}
\end{figure*}

\paragraph{VGG.} serves as a strong baseline for class sensitivity analysis due to its simple, sequential architecture. As the attribution layer activates broadly across many channels, we expect higher top-1/top-2 agreement. The distributions in ~\autoref{fig:vgg_agreement_distributions} show that most methods yield class-insensitive explanations, with significant overlap above the 50\% threshold.

\begin{figure*}[!ht]
    \centering
    \textbf{Feature Agreement Score Distributions for ResNet}\\[1em]
    \begin{minipage}[b]{0.30\textwidth}
        \centering
        \includegraphics[width=\linewidth]{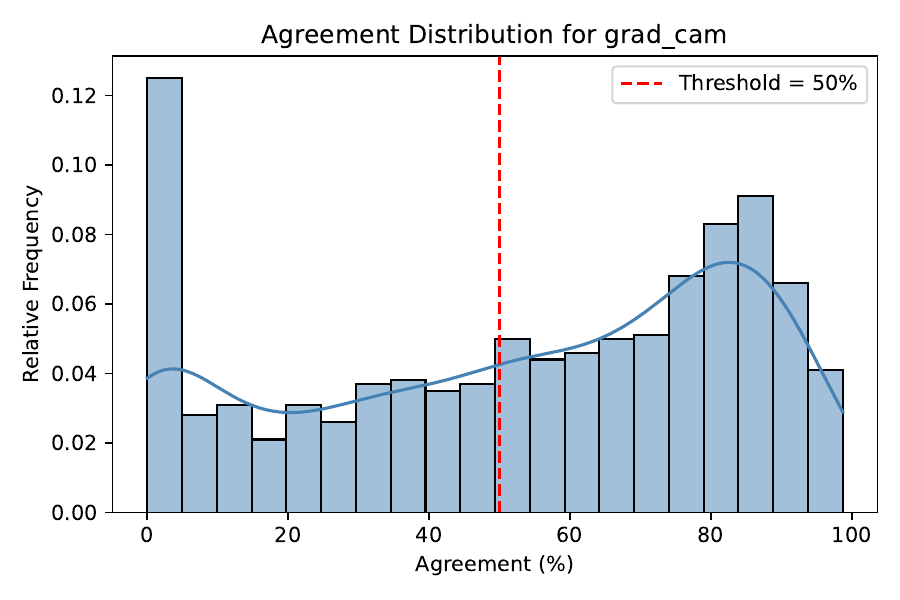}
        \\ \texttt{Grad-CAM}
    \end{minipage}
    \hfill
    \begin{minipage}[b]{0.30\textwidth}
        \centering
        \includegraphics[width=\linewidth]{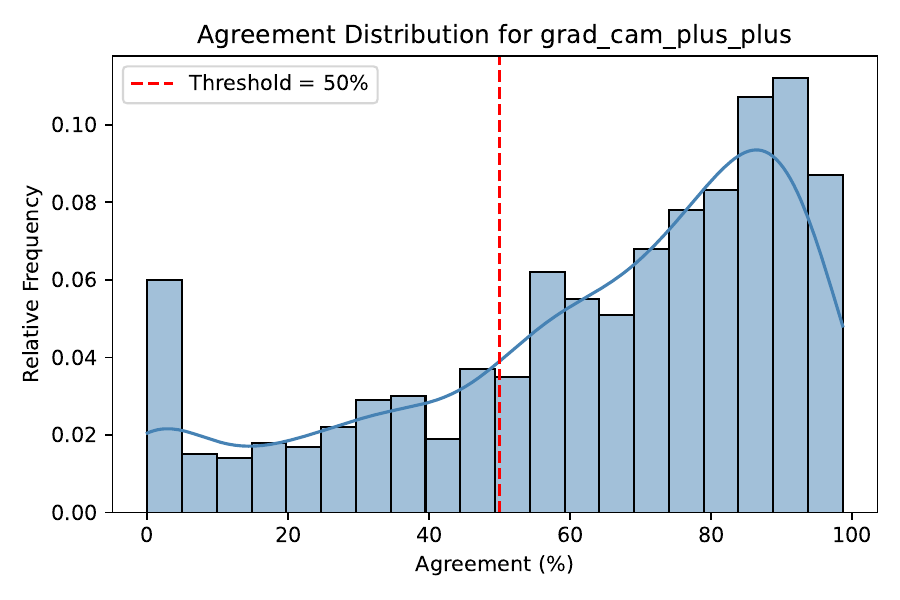}
        \\ \texttt{Grad-CAM++}
    \end{minipage}
    \hfill
    \begin{minipage}[b]{0.30\textwidth}
        \centering
        \includegraphics[width=\linewidth]{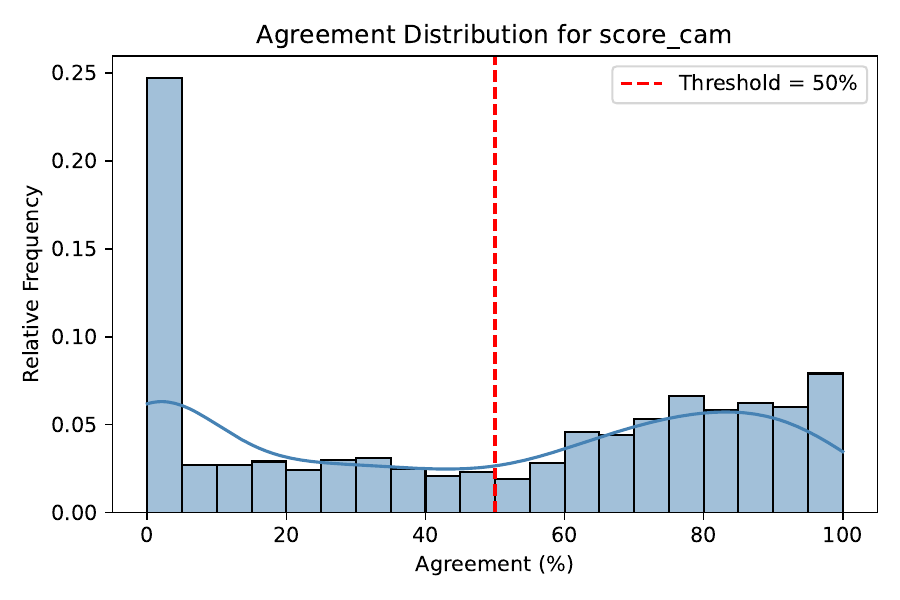}
        \\ \texttt{ScoreCAM}
    \end{minipage}

    \vspace{1em}

    \begin{minipage}[b]{0.30\textwidth}
        \centering
        \includegraphics[width=\linewidth]{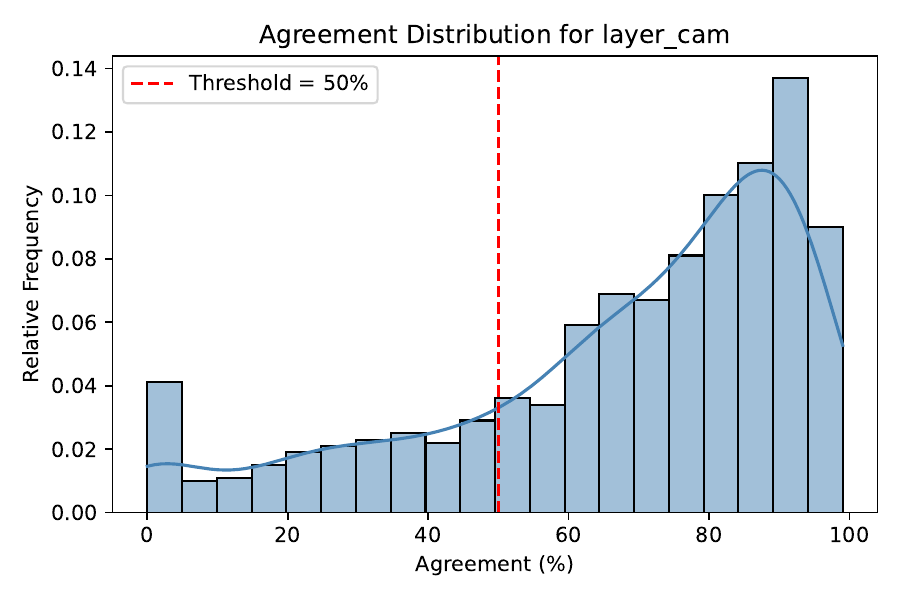}
        \\ \texttt{LayerCAM}
    \end{minipage}
    \hfill
    \begin{minipage}[b]{0.30\textwidth}
        \centering
        \includegraphics[width=\linewidth]{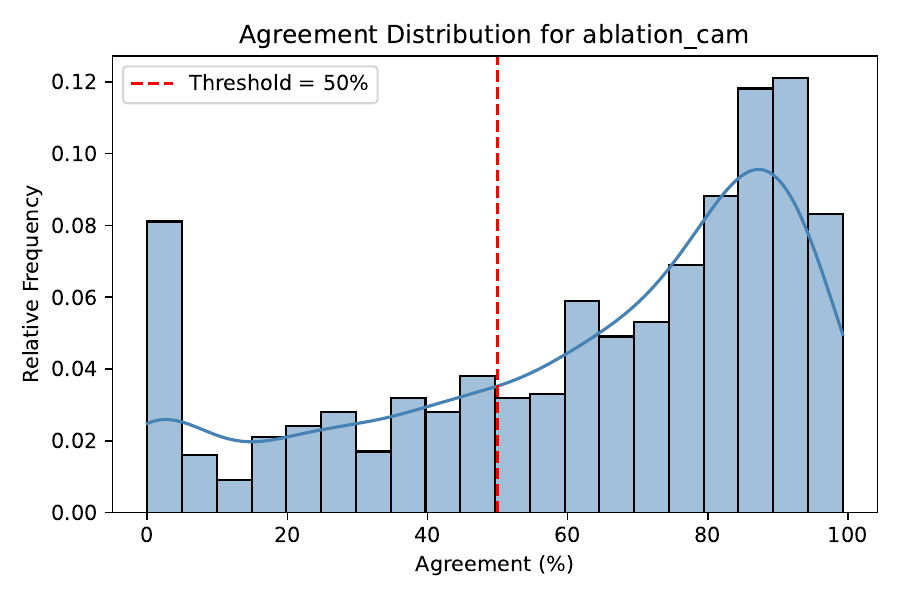}
        \\ \texttt{AblationCAM}
    \end{minipage}
    \hfill
    \begin{minipage}[b]{0.30\textwidth}
        \centering
        \includegraphics[width=\linewidth]{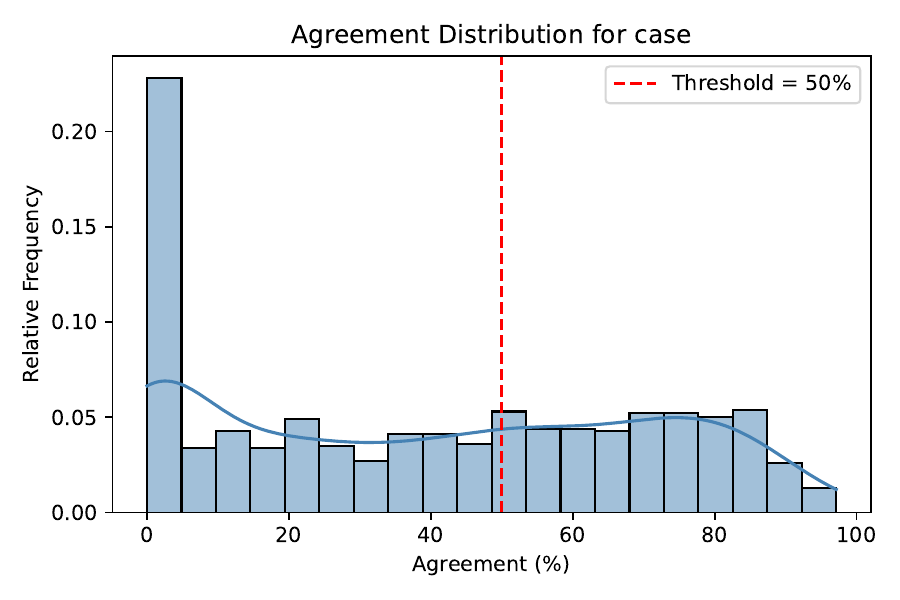}
        \\ \texttt{CASE}
    \end{minipage}

    \caption{
        Agreement score distributions for all saliency methods (ResNet). Each histogram shows the distribution of top-5\% feature overlap between saliency maps generated for top-1 and top-2 class labels. The red dashed line denotes the 50\% threshold used in our Wilcoxon test for class sensitivity.
    }
    \label{fig:resnet_agreement_distributions}
\end{figure*}

\paragraph{ResNet-50.} ResNet introduces residual connections that promote feature reuse and broader spatial coverage. (~\autoref{fig:resnet_agreement_distributions}) reveals that many saliency methods still produce class-indistinct maps, with distributions often centered near or above the 50\% threshold. CASE, however, consistently shifts agreement lower, reflecting stronger separation of class-relevant features. 

\begin{figure*}[!ht]
    \centering
    \textbf{Feature Agreement Score Distributions for ConvNeXt}\\[1em]
    \begin{minipage}[b]{0.30\textwidth}
        \centering
        \includegraphics[width=\linewidth]{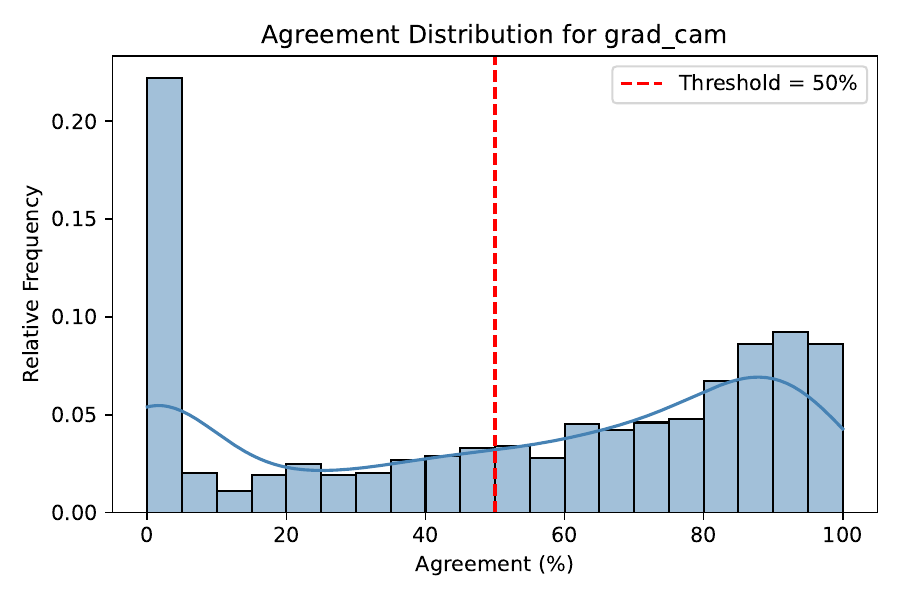}
        \\ \texttt{Grad-CAM}
    \end{minipage}
    \hfill
    \begin{minipage}[b]{0.30\textwidth}
        \centering
        \includegraphics[width=\linewidth]{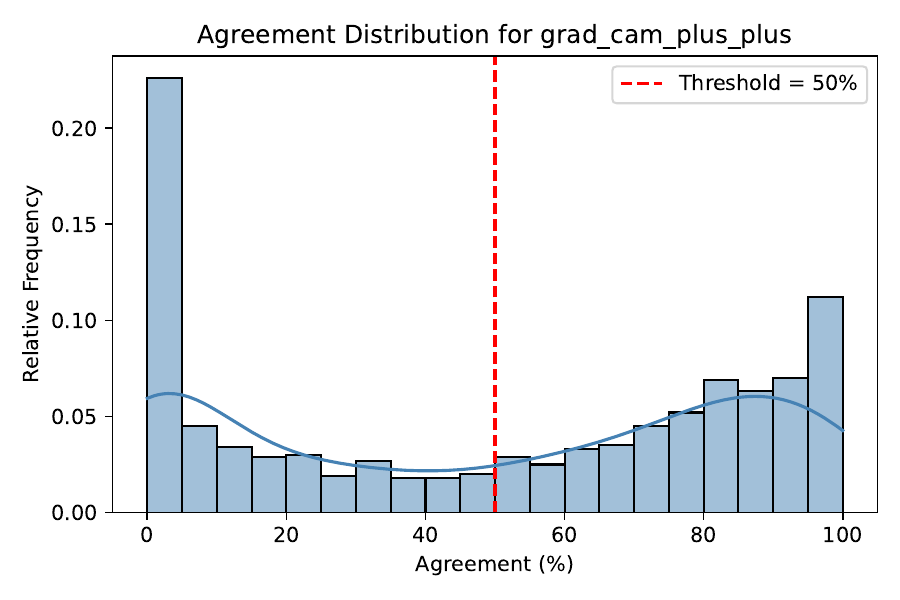}
        \\ \texttt{Grad-CAM++}
    \end{minipage}
    \hfill
    \begin{minipage}[b]{0.30\textwidth}
        \centering
        \includegraphics[width=\linewidth]{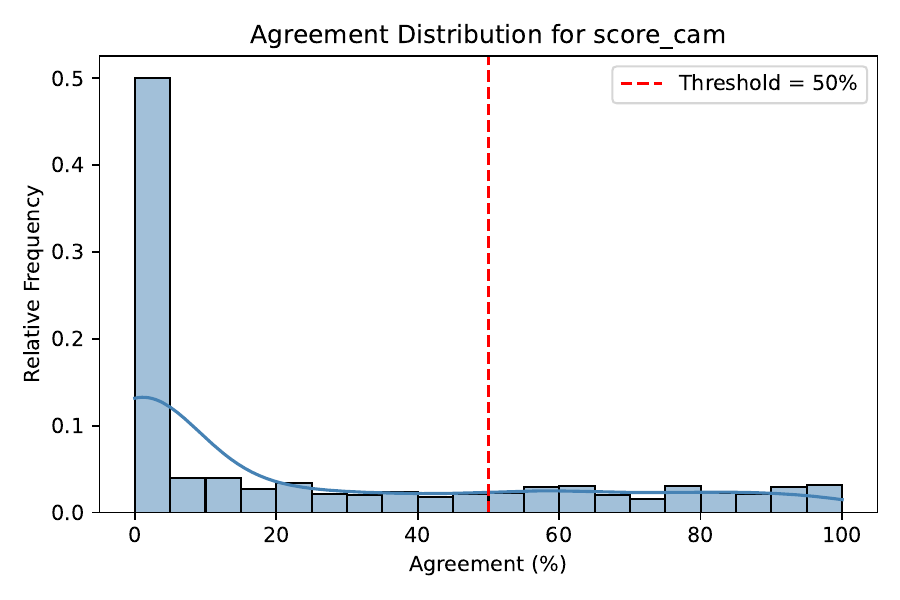}
        \\ \texttt{ScoreCAM}
    \end{minipage}

    \vspace{1em}

    \begin{minipage}[b]{0.30\textwidth}
        \centering
        \includegraphics[width=\linewidth]{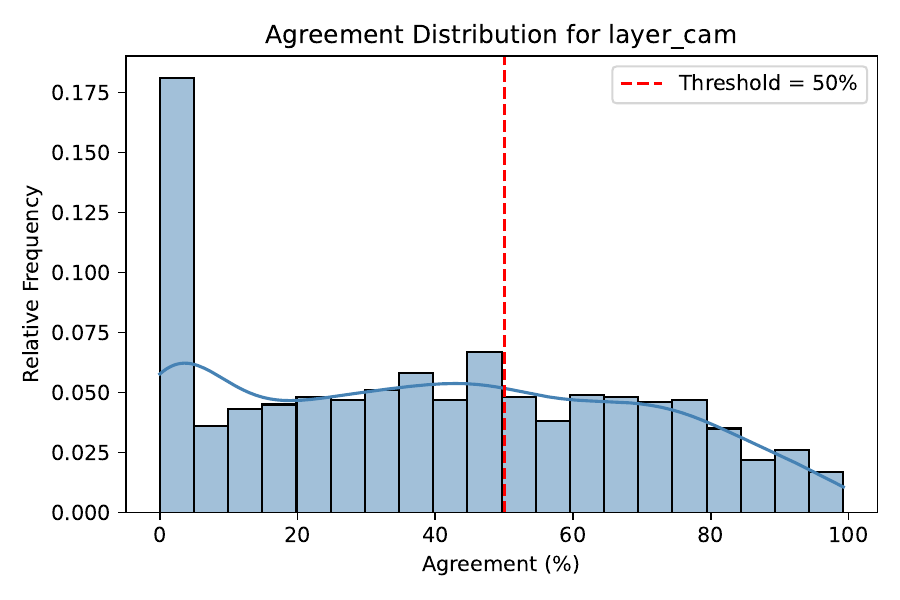}
        \\ \texttt{LayerCAM}
    \end{minipage}
    \hfill
    \begin{minipage}[b]{0.30\textwidth}
        \centering
        \includegraphics[width=\linewidth]{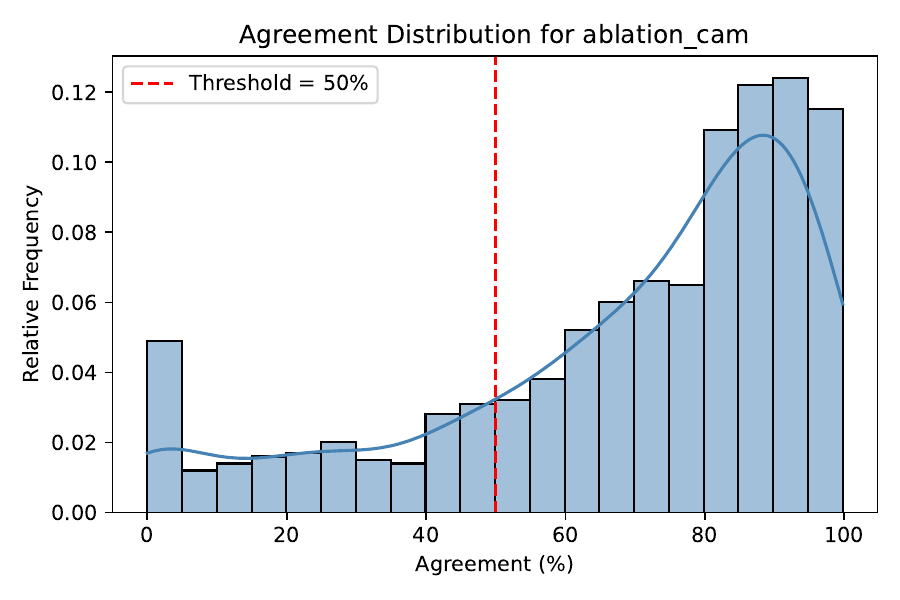}
        \\ \texttt{AblationCAM}
    \end{minipage}
    \hfill
    \begin{minipage}[b]{0.30\textwidth}
        \centering
        \includegraphics[width=\linewidth]{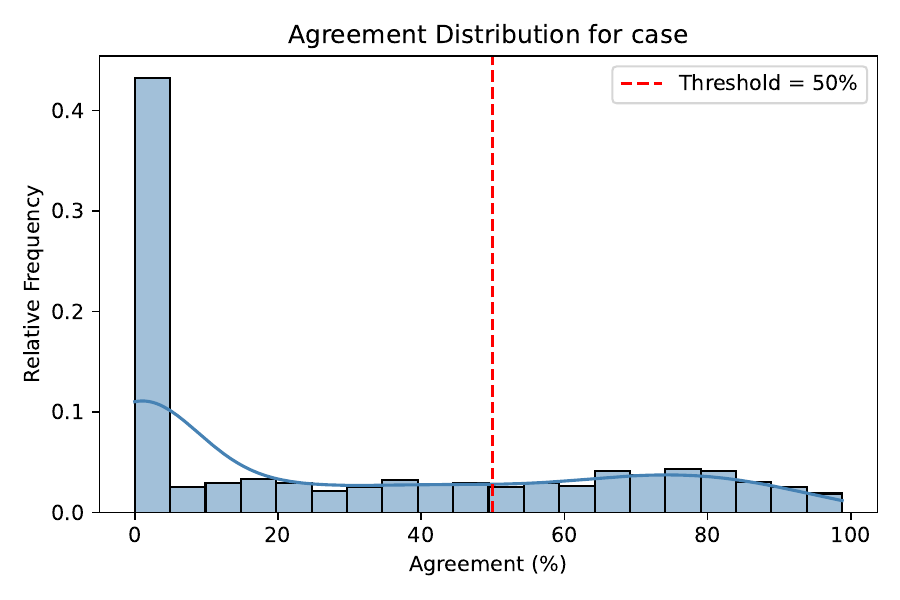}
        \\ \texttt{CASE}
    \end{minipage}

    \caption{
        Agreement score distributions for all saliency methods (ConvNeXt). Each histogram shows the distribution of top-5\% feature overlap between saliency maps generated for top-1 and top-2 class labels. The red dashed line denotes the 50\% threshold used in our Wilcoxon test for class sensitivity.
    }
    \label{fig:convnext_agreement_distributions}
\end{figure*}

\paragraph{ConvNeXt} ConvNeXt offers a modern convolutional backbone with enhanced channel mixing and normalization. Despite its design, several methods exhibit high agreement scores across class predictions (see ~\autoref{fig:convnext_agreement_distributions}), indicating limited class specificity. CASE maintains a clear advantage, with distributions skewed leftward below the threshold.

\begin{figure*}[!ht]
    \centering
    \textbf{Feature Agreement Score Distributions for DenseNet (Conv)}\\[1em]
    \begin{minipage}[b]{0.30\textwidth}
        \centering
        \includegraphics[width=\linewidth]{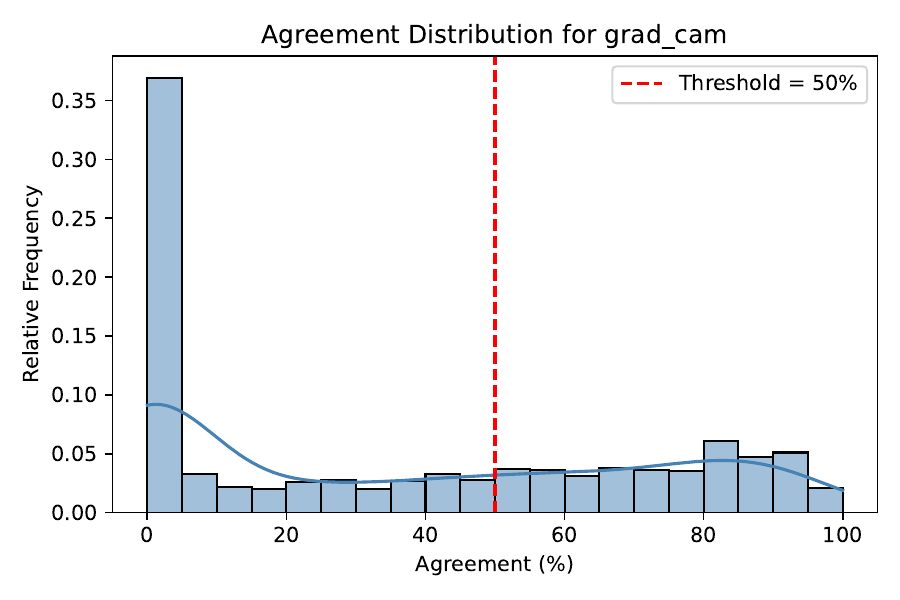}
        \\ \texttt{Grad-CAM}
    \end{minipage}
    \hfill
    \begin{minipage}[b]{0.30\textwidth}
        \centering
        \includegraphics[width=\linewidth]{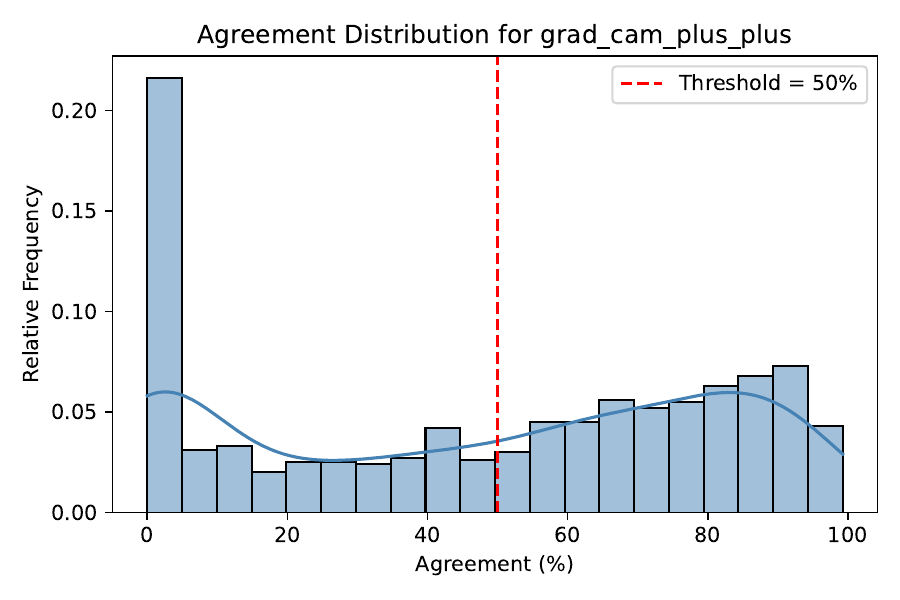}
        \\ \texttt{Grad-CAM++}
    \end{minipage}
    \hfill
    \begin{minipage}[b]{0.30\textwidth}
        \centering
        \includegraphics[width=\linewidth]{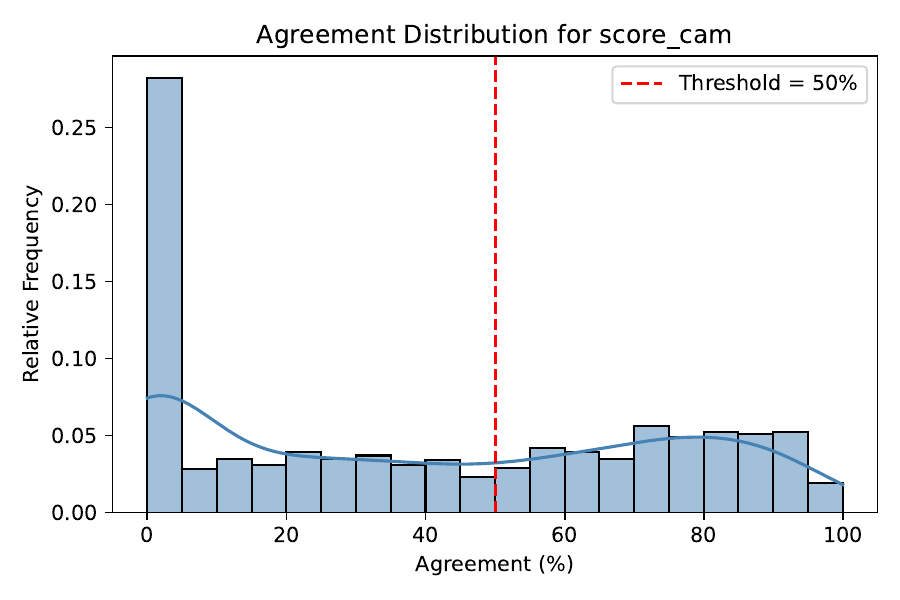}
        \\ \texttt{ScoreCAM}
    \end{minipage}

    \vspace{1em}

    \begin{minipage}[b]{0.30\textwidth}
        \centering
        \includegraphics[width=\linewidth]{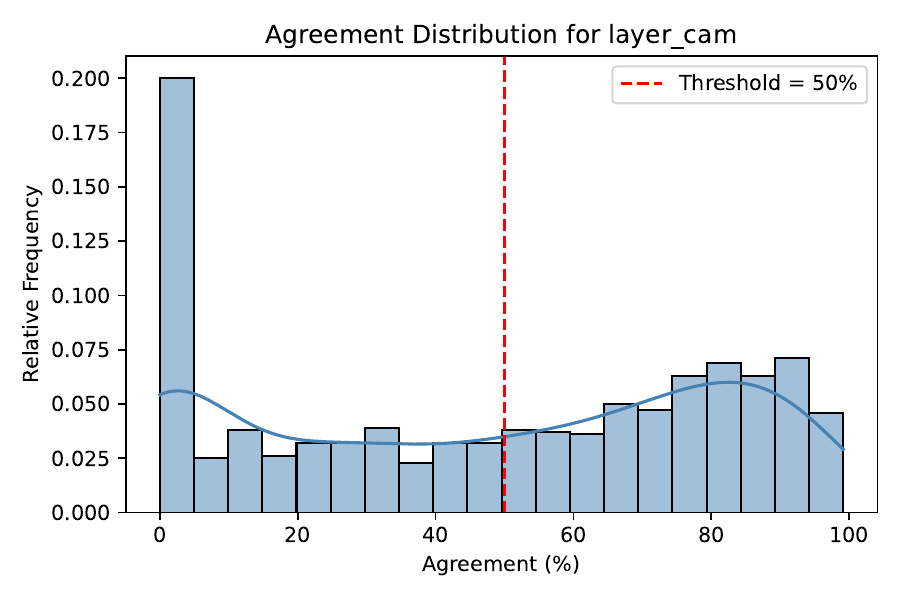}
        \\ \texttt{LayerCAM}
    \end{minipage}
    \hfill
    \begin{minipage}[b]{0.30\textwidth}
        \centering
        \includegraphics[width=\linewidth]{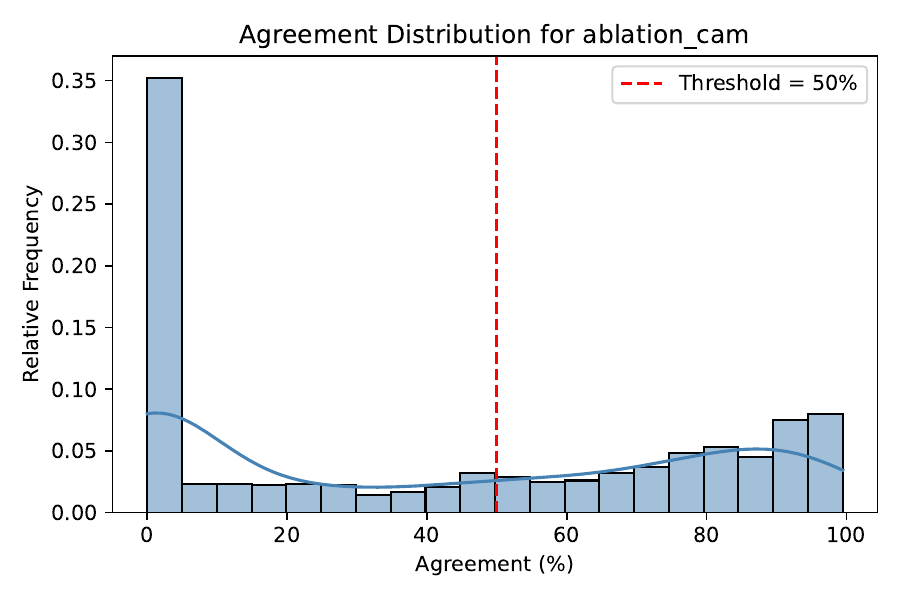}
        \\ \texttt{AblationCAM}
    \end{minipage}
    \hfill
    \begin{minipage}[b]{0.30\textwidth}
        \centering
        \includegraphics[width=\linewidth]{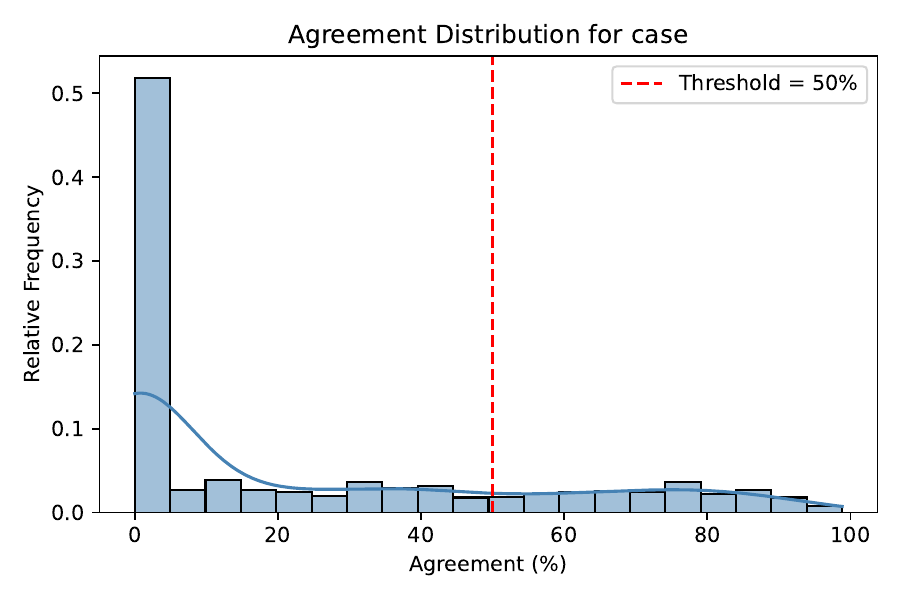}
        \\ \texttt{CASE}
    \end{minipage}

    \caption{
        Agreement score distributions for all saliency methods (DenseNet (Conv)). Each histogram shows the distribution of top-5\% feature overlap between saliency maps generated for top-1 and top-2 class labels. The red dashed line denotes the 50\% threshold used in our Wilcoxon test for class sensitivity.
    }
    \label{fig:densenet_conv_agreement_distributions}
\end{figure*}

\paragraph{DenseNet-201 (\texttt{features.denseblock4.denselayer32.conv2}).} This is the final convolutional layer used in our main experiments. As detailed in Section~\ref{sec:results_channel_sparsity}, it is extremely sparse yet highly class-discriminative. The agreement distributions here (see ~\autoref{fig:densenet_conv_agreement_distributions})
confirm that most methods produce low-overlap explanations, consistent with the high class sensitivity reported in RQ1.

\begin{figure*}[!ht]
    \centering
    \textbf{Feature Agreement Score Distributions for DenseNet (norm5)}\\[1em]
    \begin{minipage}[b]{0.30\textwidth}
        \centering
        \includegraphics[width=\linewidth]{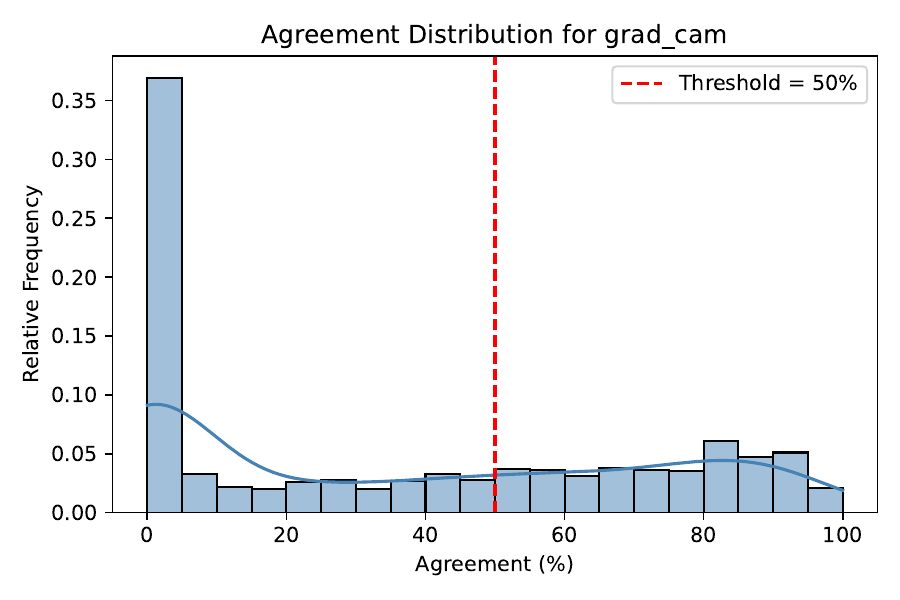}
        \\ \texttt{Grad-CAM}
    \end{minipage}
    \hfill
    \begin{minipage}[b]{0.30\textwidth}
        \centering
        \includegraphics[width=\linewidth]{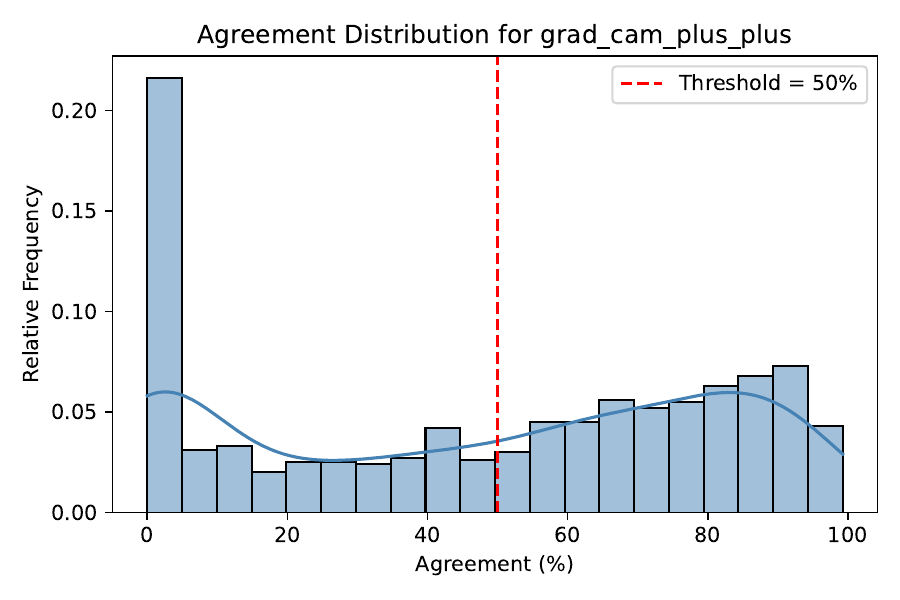}
        \\ \texttt{Grad-CAM++}
    \end{minipage}
    \hfill
    \begin{minipage}[b]{0.30\textwidth}
        \centering
        \includegraphics[width=\linewidth]{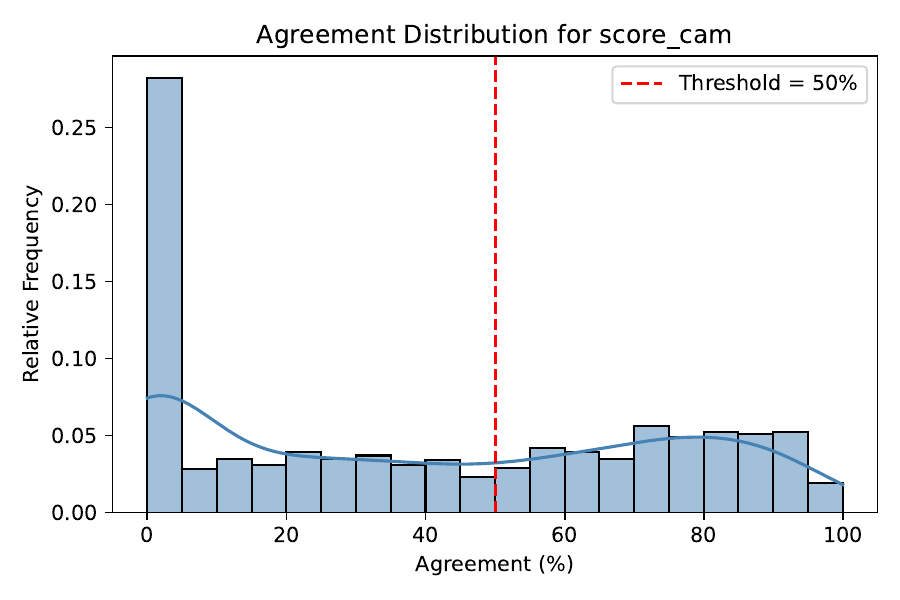}
        \\ \texttt{ScoreCAM}
    \end{minipage}

    \vspace{1em}

    \begin{minipage}[b]{0.30\textwidth}
        \centering
        \includegraphics[width=\linewidth]{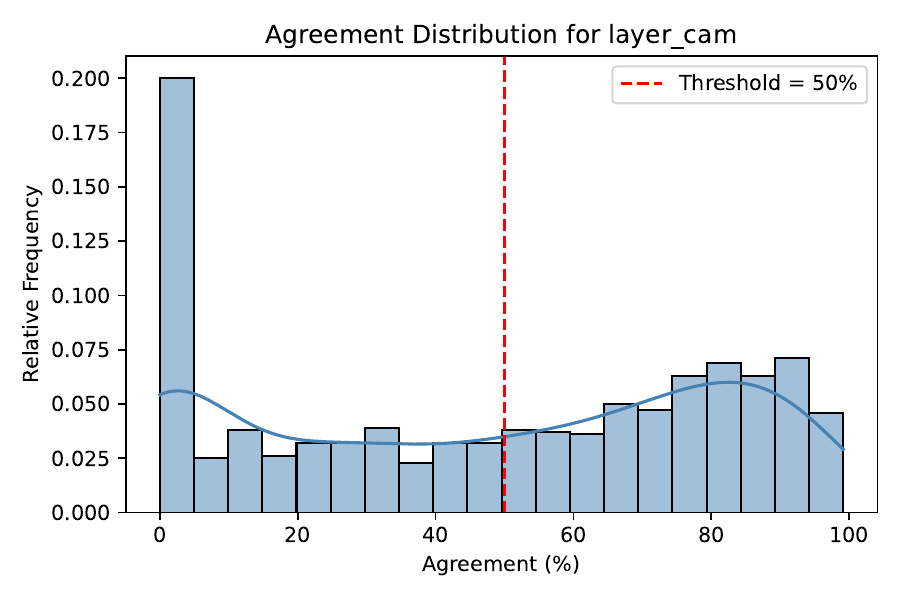}
        \\ \texttt{LayerCAM}
    \end{minipage}
    \hfill
    \begin{minipage}[b]{0.30\textwidth}
        \centering
        \includegraphics[width=\linewidth]{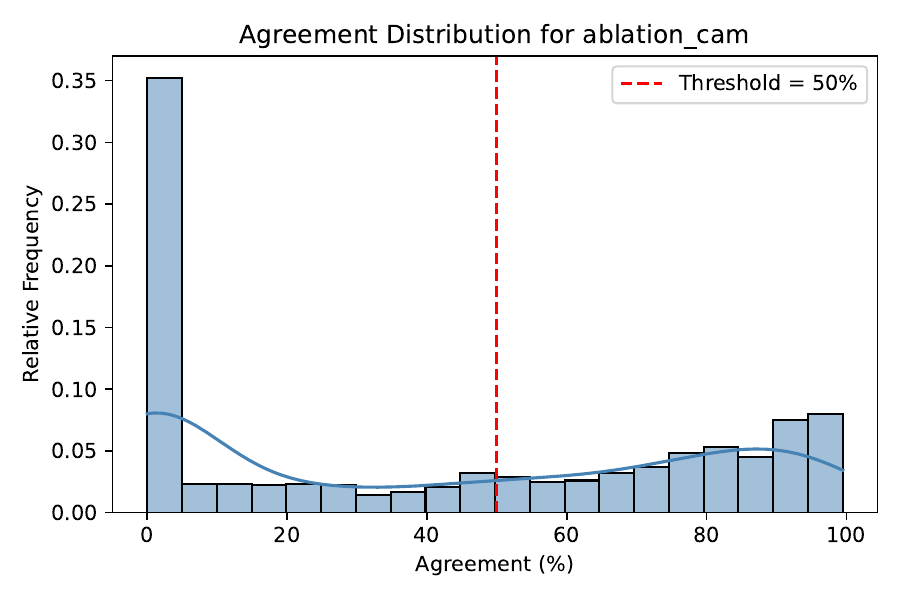}
        \\ \texttt{AblationCAM}
    \end{minipage}
    \hfill
    \begin{minipage}[b]{0.30\textwidth}
        \centering
        \includegraphics[width=\linewidth]{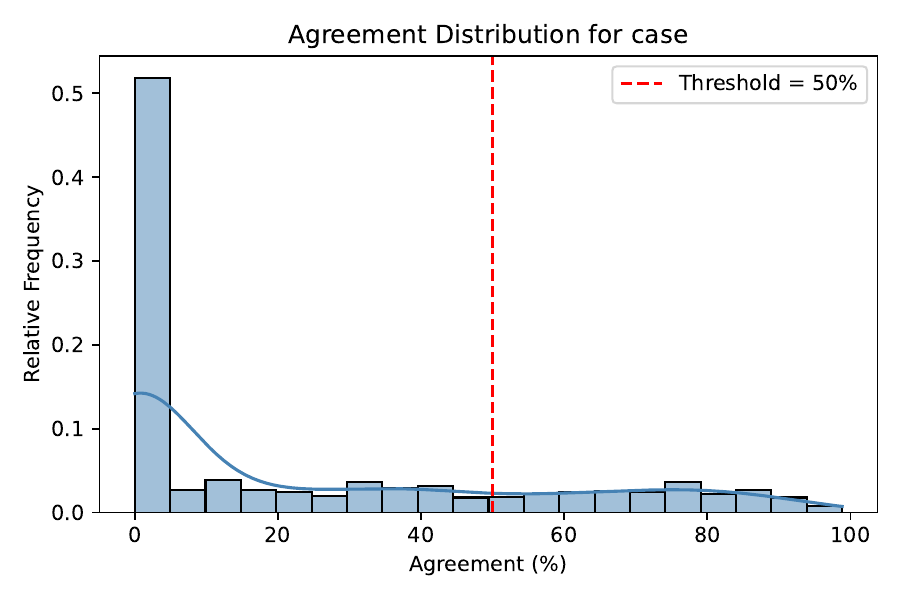}
        \\ \texttt{CASE}
    \end{minipage}

    \caption{
        Agreement score distributions for all saliency methods (DenseNet (Norm5)). Each histogram shows the distribution of top-5\% feature overlap between saliency maps generated for top-1 and top-2 class labels. The red dashed line denotes the 50\% threshold used in our Wilcoxon test for class sensitivity.
    }
    \label{fig:densenet_norm_agreement_distributions}
\end{figure*}

\paragraph{DenseNet-201 (\texttt{features.norm5}).} This final normalization layer aggregates upstream activations and is located immediately before the classifier. As discussed in our ablation study, this layer activates nearly all channels uniformly. The agreement distributions (~\autoref{fig:densenet_norm_agreement_distributions}) reveal that most methods fail to produce class-distinct saliency maps in this setting, highlighting the importance of attribution layer selection.

\end{document}